\documentclass[journal]{IEEEtran}

\ifCLASSINFOpdf
  \usepackage[pdftex]{graphicx}
  \graphicspath{{./imgs/}}
  \DeclareGraphicsExtensions{.pdf,.jpeg,.png}
\else
  \usepackage[dvips]{graphicx}
  \graphicspath{{./imgs/}}
  \DeclareGraphicsExtensions{.eps}
\fi

\usepackage{cite}
\usepackage{xcolor} 
\usepackage[colorlinks,
            linkcolor=blue,
            anchorcolor=blue,
            citecolor=blue]{hyperref}

\usepackage{multirow}
\usepackage{graphicx} 
\usepackage{makecell}
\usepackage{array}
\usepackage{subfig}
\usepackage{amsmath}
\usepackage{amssymb}
\usepackage{booktabs} 
\usepackage{subfig}  

\usepackage{caption}

\usepackage{soul, color, xcolor}

\soulregister{\cite}7 
\soulregister{\citep}7 
\soulregister{\citet}7 
\soulregister{\ref}7 
\soulregister{\pageref}7 

\begin{document}
%
\title{Domain Adaptive SAR Wake Detection: Leveraging Similarity Filtering and Memory Guidance}
%
%
%

\author{He~Gao,
        Baoxiang~Huang,~\IEEEmembership{Senior~Member,~IEEE,}
        Milena~Radenkovic,~\IEEEmembership{Member,~IEEE,}
        Borui~Li,
        and~Ge~Chen,~\IEEEmembership{Senior~Member,~IEEE}
\thanks{This research was financially supported by the National Natural Science Foundation of China No.42276203, 42530404), the Laoshan Laboratory (No.LSKI202204302) and the Discipline Cluster Research Project of Qingdao University "Deep mining, and intelligent prediction of multimodal big data for marine ecological disasters"(Grant No.XT2024101).(Corresponding author: Baoxiang Huang) } 
\thanks{He Gao and Borui Li is with the College of Computer Science and Technology, Qingdao University, Qingdao 266071, China (e-mail: ghsticker@163.com, liborui@qdu.edu.cn).}
\thanks{Baoxiang Huang is with the College of Computer Science and Technology, Qingdao University, Qingdao 266071, China, and also with the Laboratory for Regional Oceanography and Numerical Modeling, Qingdao Marine Science and Technology Center, Qingdao 266100, China (e-mail: hbx3726@163.com).}
\thanks{Milena Radenkovic is with the School of Computer Science and Information Technology, The University of Nottingham, NG8 1BB Nottingham, U.K. (e-mail: mvr@cs.nott.ac.uk).}
\thanks{Ge Chen is with the State Key Laboratory of Physical Oceanography, Department of Marine Technology, Ocean University of China, Qingdao 266100, China, and also with the Laboratory for Regional Oceanography and Numerical Modeling, Department of Ocean Big Data and Prediction, Laoshan Laboratory, Qingdao 266100, China (e-mail: gechen@ouc.edu.cn).}
}
%
%

\markboth{Journal of \LaTeX\ Class Files,~Vol.~13, No.~9, September~2025}%
{Shell \MakeLowercase{\textit{et al.}}: Bare Demo of IEEEtran.cls for Journals}
%



\maketitle

\begin{abstract}
Synthetic Aperture Radar (SAR), with its all-weather, all-day, and wide-area observation capabilities, has become an important remote sensing data source in the field of ship wake detection, widely applied in maritime surveillance, ship tracking, and military defense. However, the complex imaging mechanism of SAR leads to wake features that are abstract, weak, and easily disturbed by noise, making it difficult to achieve accurate and large-scale annotations. In contrast, optical images possess more intuitive visual features and are relatively easier to annotate, but directly transferring models trained on optical images to SAR images often results in performance degradation. To address this cross-modal domain adaptation challenge, we propose a Similarity-Guided and Memory-Guided Domain Adaptation (termed SimMemDA) framework for unsupervised domain adaptive ship wake detection via instance-level feature similarity filtering and feature memory guidance. Specifically, we design a wake structure-preserving style transfer model, WakeGAN, in which we introduce frequency selection, detail enhancement, and structure preserving modules into the generator. Combined with spectral preservation and cyclic spectral consistency losses, this model imposes dual constraints on geometric structure and physical texture, thereby narrowing the domain gap at the input level. Furthermore, through instance-level feature distribution modeling, we select source domain samples whose feature distributions are more similar to those of the target domain, reducing the risk of negative transfer during model training. At the same time, we construct a memory-guided geometric-aware pseudo-label calibration mechanism, which leverages neighborhood feature consistency and wake-line geometric priors to enhance pseudo-label reliability. Finally, we adopt a region-mixing training strategy, integrating real annotations from the source domain with refined high-quality pseudo-labels from the target domain, thereby further improving the generalization capability of the model. Experimental results demonstrate that the proposed SimMemDA method improves both accuracy and robustness in cross-modal ship wake detection tasks, validating the effectiveness and feasibility of the approach.

\end{abstract}


%
\IEEEpeerreviewmaketitle

\section{Introduction}

\IEEEPARstart{S}{hip} wake detection in Synthetic Aperture Radar (SAR) imagery plays a crucial role in maritime surveillance, vessel tracking, and military defense applications\cite{ding2023towards},\cite{guan2023method},\cite{xin2023ship},\cite{xu2024wake2wake}. Ship wakes, as the most distinctive signatures of moving vessels in remote sensing images, typically extend several tens of kilometers and encapsulate essential information, including vessel structure, velocity, and geometric characteristics. Compared to methods solely detecting ships, simultaneous detection of vessels and their wakes proves more practical and informative. The underlying imaging mechanism for SAR-based ship wake detection involves the interaction between ship-generated wakes and gravity-capillary waves. This interaction modulates sea surface roughness, leading to alterations in radar backscatter signals and resulting in clearly identifiable wake patterns within SAR imagery \cite{vesecky1982observation}. SAR technology, independent of natural illumination sources, enables high-resolution, extensive-area surveillance of maritime environments under diverse meteorological conditions (e.g., cloud coverage, precipitation, fog) and varying lighting scenarios (both daytime and nighttime). Consequently, SAR payloads have emerged as essential data sources for effective and reliable ship wake detection.

In the study of ship wake detection in SAR imagery, existing approaches can be broadly categorized into two groups: traditional hand-crafted feature-based methods and deep learning-based methods. Traditional methods primarily rely on classical image processing techniques, such as edge detection, Radon transform, and constant false alarm rate (CFAR) detection  to capture the linear structural characteristics of wakes \cite{biondi2018polarimetric}, \cite{jiaqiu2011novel}. These methods can achieve certain effectiveness under controlled environmental conditions. However, they are highly sensitive to complex sea states and the inherent speckle noise of SAR images. As a result, they often require additional filtering and preprocessing steps to reduce false alarms, which limits their robustness in real-world ocean environments. In contrast, deep learning-based approaches leverage convolutional neural networks and object detection frameworks \cite{kang2019ship}, \cite{ding2023towards}, \cite{xu2024wake2wake} to automatically learn high-level semantic representations of wakes, exhibiting significant advantages in detection accuracy and adaptability. Nevertheless, such methods heavily depend on large-scale annotated datasets. SAR wake data, however, are not only scarce in publicly available resources but also costly and labor-intensive to annotate, which severely restricts their scalability in practical applications. Against this backdrop, reducing reliance on large-scale SAR annotations has become a core challenge in this field. Given that optical wake data are relatively easier to acquire and annotate, cross-domain detection provides a feasible pathway to alleviate the scarcity of SAR training data.

With the advancement of satellite remote sensing technologies, the utilization of multi-payload remote sensing imagery for ship wake detection has become increasingly practical. Optical imagery offers a complementary perspective due to its reliance on visual attributes such as color, brightness, texture, and shape, enabling clear delineation of wake structures. This significantly facilitates data annotation, thereby making the development of extensive annotated datasets feasible \cite{liu2018ship}. Nevertheless, the quality of optical imagery is frequently compromised by environmental conditions, including cloud coverage and limited nighttime imaging capabilities, thus constraining its continuous applicability in maritime surveillance. In contrast, SAR imagery, characterized by all-weather and day-and-night operational capabilities along with certain penetration abilities, demonstrates distinctive advantages in ship wake monitoring. Therefore, the integration of complementary information from both optical and SAR imagery is expected to substantially enhance the robustness and adaptability of ship wake detection systems. However, the fundamental differences in the imaging mechanisms result in significant discrepancies in the representation of ship wakes between optical and SAR imagery.
\begin{figure}[!t]
  \centering
  \subfloat[]{%
    \includegraphics[width=0.48\columnwidth]{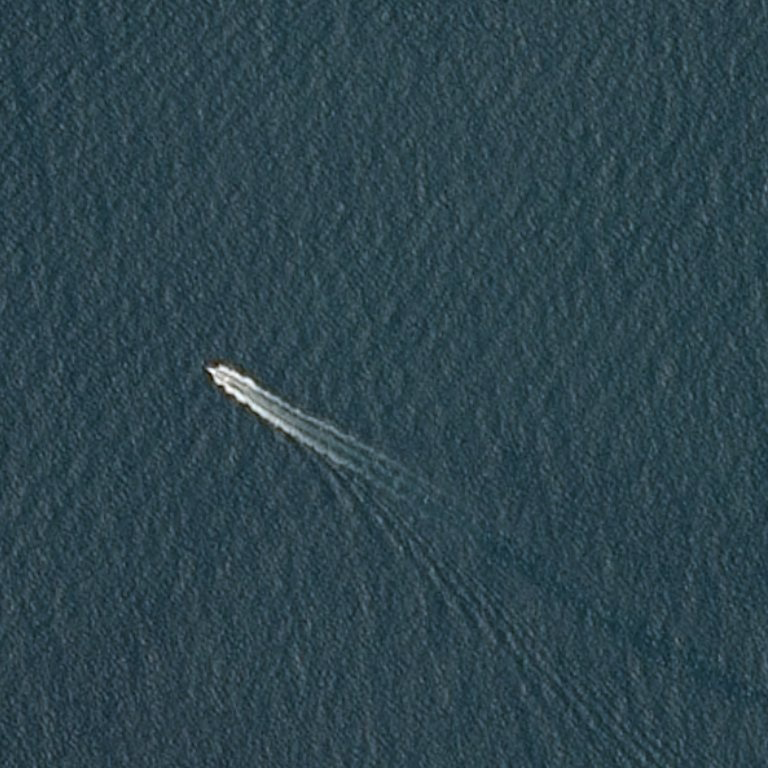}%
    \label{fig:optical}
  }
  \hfill
  \subfloat[]{%
    \includegraphics[width=0.48\columnwidth]{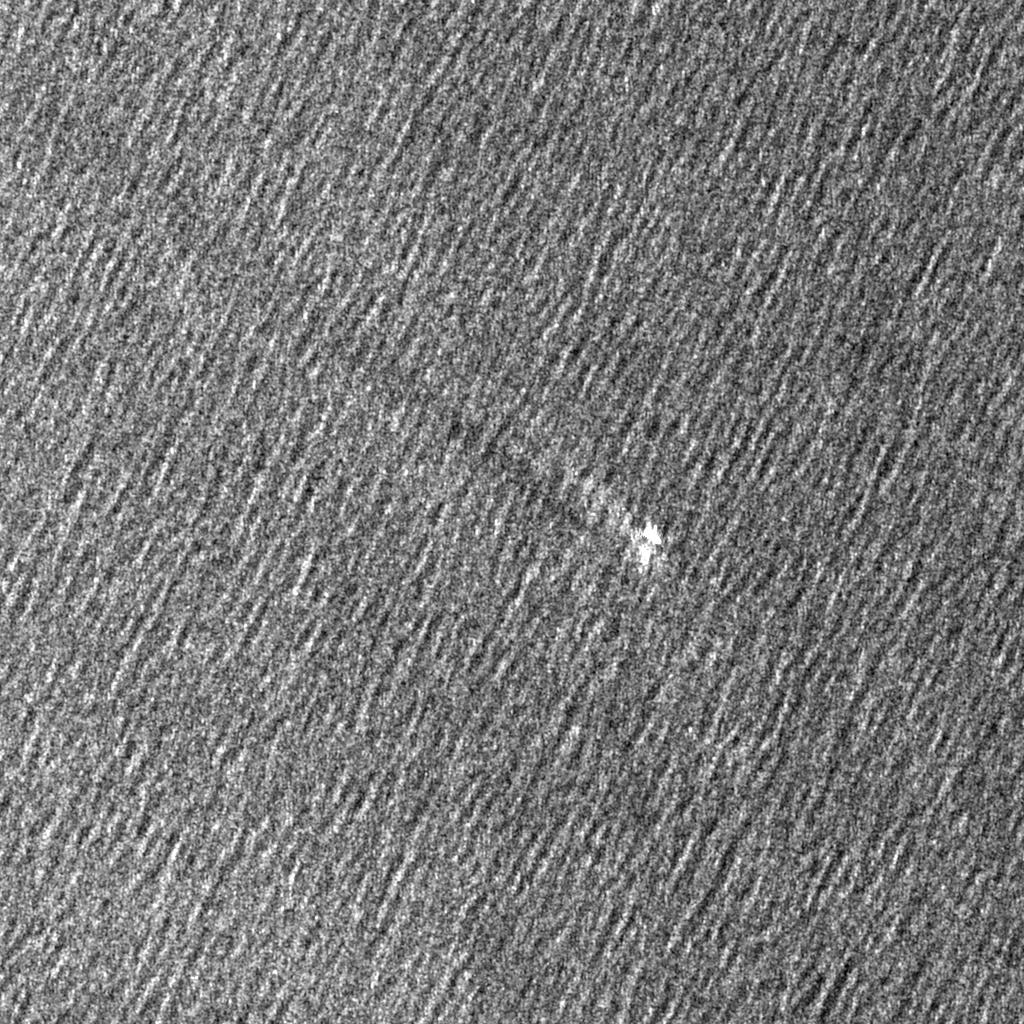}%
    \label{fig:sar}
  }
  \caption{Visual comparison of optical and SAR ship wakes. (a) Ship wake image captured by optical sensors; (b) Ship wake image acquired using SAR sensors.}
  \label{fig:oprical_sar}
\end{figure}

\begin{figure}[!t]
  \centering
  \subfloat[]{%
    \includegraphics[width=0.48\columnwidth]{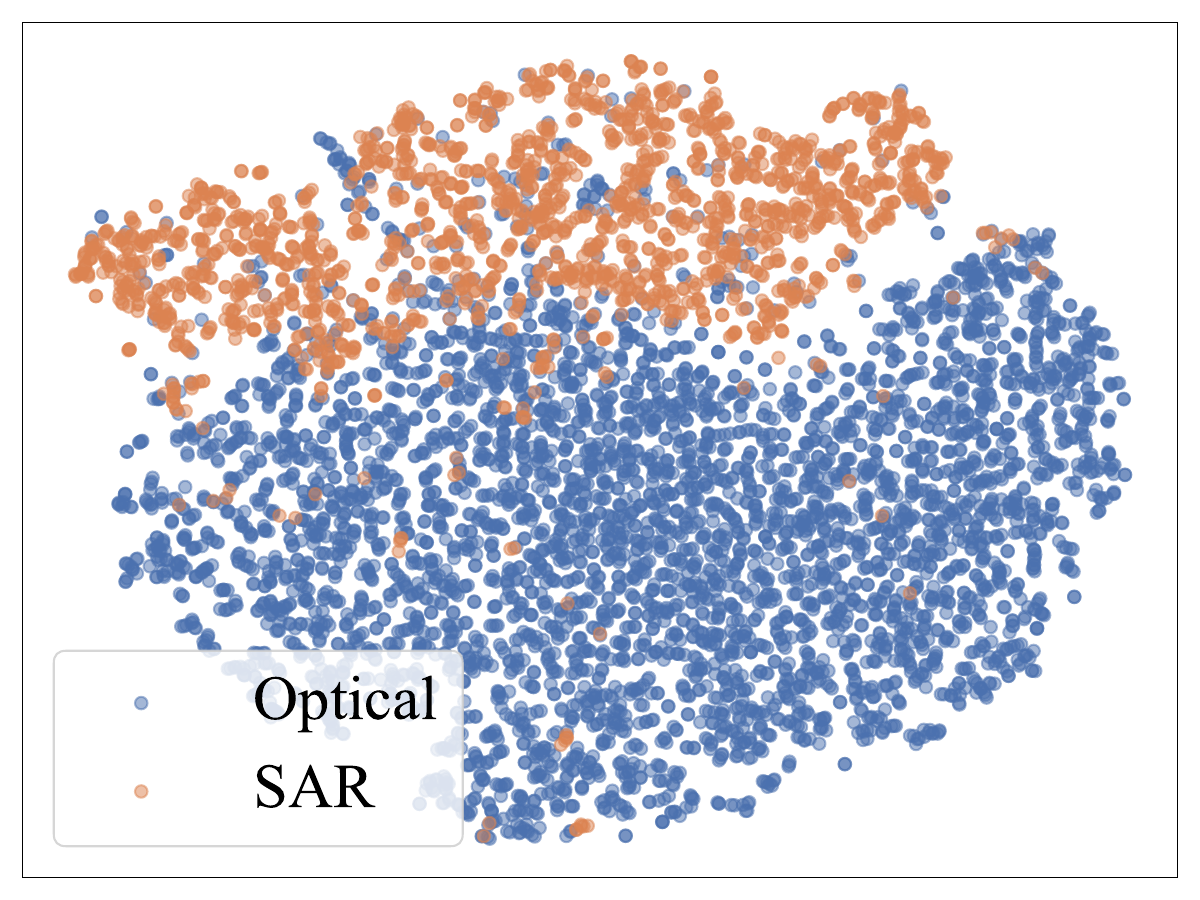}%
    \label{fig:tsne}
  }
  \hfill
  \subfloat[]{%
    \includegraphics[width=0.48\columnwidth]{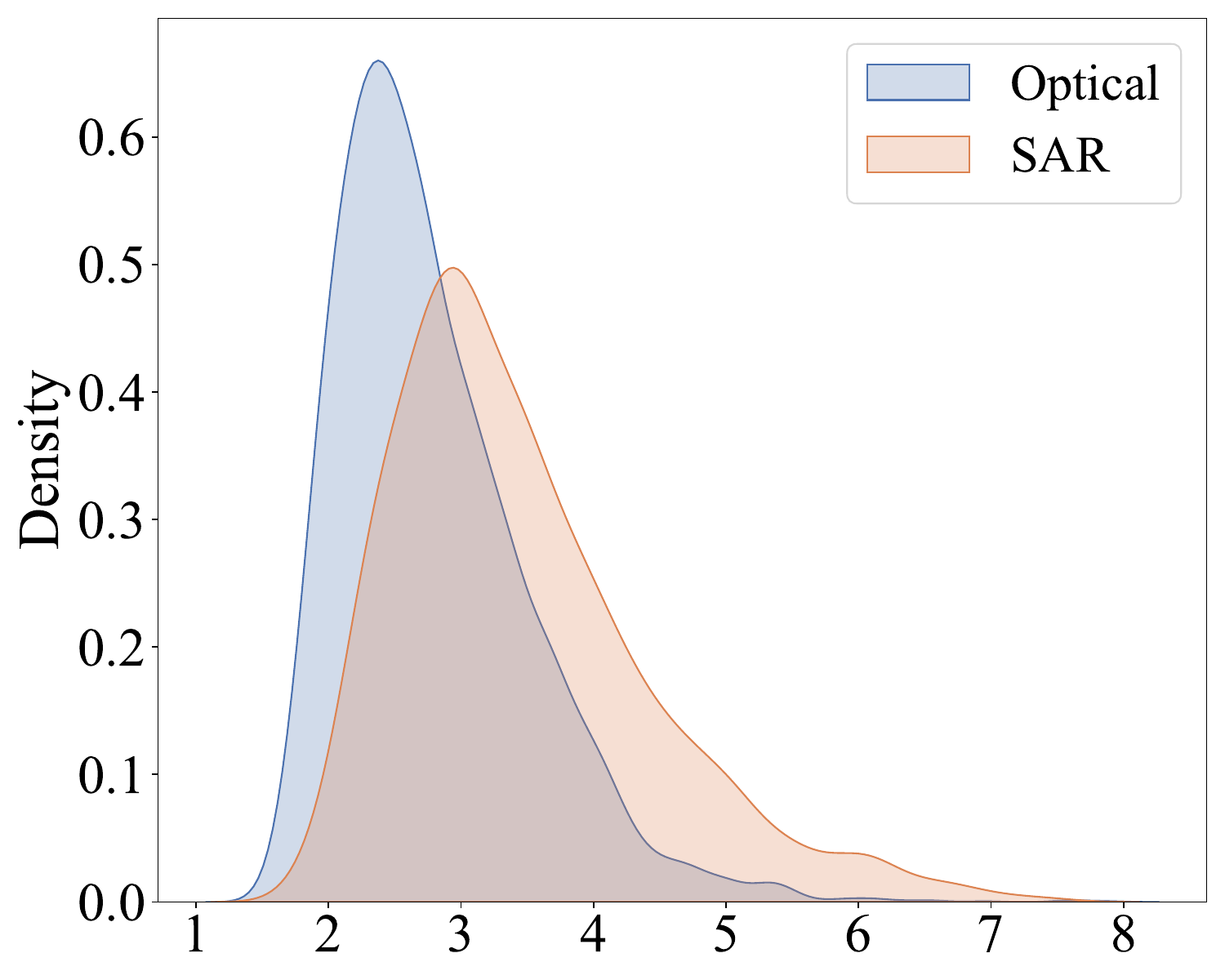}%
    \label{fig:kde}
  }

  \caption{(a) Comparison of feature distributions between optical ship wakes and SAR ship wakes, obtained using t-SNE dimensionality reduction; (b) Intra-class distributions of wakes from optical and SAR images, where the $x$-axis represents the Euclidean distance from the data to the center of source domain.}
  \label{fig:distribution}
\end{figure}

As depicted in Fig. \ref{fig:oprical_sar}, optical imagery explicitly exhibits detailed wake structures, whereas wakes in SAR imagery typically appear abstract, heavily influenced by noise and geometric distortions inherent to SAR imaging. Consequently, direct transfer of a detection model trained on optical imagery to SAR imagery generally leads to a notable performance reduction, complicating the direct application of optical imagery-derived annotated datasets to SAR-based wake detection tasks. Further analysis of feature distributions between optical and SAR imagery highlights pronounced distribution shifts within the feature space, as illustrated in Fig. \ref{fig:distribution}. To effectively capitalize on the complementary strengths of these two imagery types, Unsupervised Domain Adaptation (UDA) presents a viable solution. UDA facilitates knowledge transfer from well-annotated optical imagery (source domain) to unlabeled SAR imagery (target domain), circumventing the need for annotated SAR data. Ideally, training on richly annotated optical datasets could significantly enhance wake detection capabilities in SAR imagery. Recently, UDA methodologies have been extensively explored in object detection applications, broadly classified into four primary categories: Domain alignment methods\cite{chen2018domain},\cite{hsu2020every},\cite{li2022sigma},\cite{zhao2022task},\cite{li2022scan++},\cite{li2022scan} align features across various abstraction levels, minimizing domain discrepancies and enhancing generalization capabilities. Pseudo-labeling methods\cite{li2021category},\cite{ramamonjison2021simrod}, \cite{mattolin2023confmix}, \cite{yang2025versatile} leverage model-generated high-confidence predictions on target imagery as pseudo-labels to facilitate self-supervised learning, although their effectiveness is highly sensitive to inter-domain variations. Auxiliary model strategies\cite{chen2022learning},\cite{he2022cross},\cite{xu2020exploring},\cite{cao2023contrastive} employ supplementary models to bolster the main detector's stability and precision during domain adaptation, thus providing robust supervisory signals. Data generation techniques\cite{kim2019diversify},\cite{wang2021afan} use generative models for style translation between source and target imagery, thereby mitigating domain shifts. While these methods have proven successful in natural image scenarios, the distinct characteristics of remote sensing imagery pose unique challenges when directly applying these approaches to remote sensing detection tasks. Recently, many studies have applied UDA methods to the remote sensing field. Shi et al. \cite{shi2022unsupervised} proposed a progressive domain adaptation strategy transferring annotated optical imagery information to unlabeled SAR imagery at pixel, feature, and prediction levels, enhancing SAR object detection accuracy. Xi et al.\cite{xi2024cromoda} combined annotated optical data with unannotated SAR data, employing a staged target domain pseudo-label generation and self-training strategy to enhance the model's ability to learn SAR image features. Chan et al. \cite{chan2024complex} introduced a cross-domain adversarial module into YOLOv5, achieving end-to-end feature alignment to counteract performance degradation due to complex sea clutter backgrounds. Zhou et al. \cite{zhou2024domain} proposed a domain adaptive approach for few-shot ship detection, capitalizing on latent similarities between optical and SAR imagery. Despite these developments, dedicated research specifically targeting ship wakes remains relatively limited. Unlike typical remote sensing objects, ship wakes present as slender and subtle structures, rendering them particularly vulnerable to noise and geometric distortions. Although some recent self-supervised approaches employ pseudo-labeling for domain adaptation, unreliable pseudo-labels frequently lead to error propagation within the target domain. Furthermore, while domain adaptation traditionally employs adversarial alignment to address domain discrepancies, practical scenarios often contain outlier samples within the source domain significantly divergent from target distributions. If inadequately managed, such anomalous samples can inject noise into the adaptation process, impairing the model's generalization performance and efficiency.

To address these challenges, we propose a \textbf{Sim}ilarity-Guided and \textbf{Mem}ory-Guided \textbf{D}omain \textbf{A}daptation (SimMemDA) framework for unsupervised domain adaptive detection via instance-level feature similarity filtering and feature memory guidance. Specifically, to alleviate feature mismatch caused by domain appearance inconsistency, we first design WakeGAN at the input level to achieve structure-preserving style transfer, thereby mitigating cross-domain pixel distribution shift. At the instance level, we introduce similarity-guided source domain filtering to eliminate samples that differ significantly from the target domain. At the supervision level, we incorporate a memory-guided geometric-aware pseudo-label calibration strategy to enhance the reliability of pseudo-labels in the target domain. Ultimately, under the joint constraints of source domain annotations and target domain pseudo-labels, the framework effectively improves the cross-domain adaptability of wake detection.

The main contributions of this work are summarized as follows:
\begin{itemize}
\item We propose a structure-preserving style transfer model, WakeGAN, which targets the dual characteristics of long-scale wake geometry and complex scattering textures. We design a frequency selection unit, a detail enhancement guide, and a structure-preserving guide, and further introduce spectral preservation loss and cyclic spectral consistency loss to achieve geometric fidelity and physical texture consistency of wakes in cross-domain transfer.
\item We design a similarity-guided source domain data filtering strategy. By statistically modeling the distribution of wakes in the target domain, we select optical wake samples that match the characteristics of SAR wakes, while discarding instances with excessive deviations, thereby reducing noise interference and negative transfer in wake domain adaptation.
\item We propose a memory-guided geometric-aware pseudo-label calibration strategy. By combining a feature-confidence memory bank with priors of wake slenderness and directional consistency, we perform neighborhood fusion and geometric-constrained correction on pseudo-labels. Furthermore, we introduce a global-instance adaptive thresholding strategy to improve the reliability of pseudo-labels and the stability of training in the target domain.
\end{itemize}

To enhance the readability of the paper, before elaborating on the methodology, we will first provide a brief explanation of the key terms involved. Memory guidance refers to constructing a feature-confidence memory bank during training, which stores the feature representations and prediction confidences of target domain samples, and then refines pseudo-labels by incorporating neighborhood retrieval and wake geometric features, thereby leveraging prior information accumulated across training stages to improve the stability and reliability of pseudo-labels. Confidence-weighted fusion in pseudo-label optimization means that instead of relying solely on the prediction result of a single sample, it combines the confidences and feature similarities of neighboring samples through weighted averaging, in order to suppress noise interference and enhance pseudo-label quality. Cross-domain alignment refers to progressively reducing the distribution discrepancy between the source and target domains through multi-level collaborative constraints at the input level, sample level, and supervision level, thereby enhancing the cross-domain generalization ability of the model.

The remainder of this paper is organized as follows. Section \ref{section_2} provides a comprehensive review of related work. Section \ref{section_3} describes the proposed method in detail. Section \ref{section_4} illustrates the experimental setup, results, and performance evaluations to validate the effectiveness of the proposed approach. Finally, Section \ref{section_5} concludes the paper with a summary and discussion of the proposed method.

\section{Related Work}\label{section_2}
\subsection{Ship Wake Detection Method}
The development of ship wake detection techniques can be broadly categorized into two distinct stages: traditional methods and deep learning methods. Traditional methods primarily rely on image processing techniques to detect prominent linear features characteristic of ship wakes. Linear feature extraction methods such as the Radon transform \cite{beylkin1985imaging} and the Hough transform \cite{ballard1981generalizing} have been extensively employed. Specifically, the Radon transform effectively detects linear wake features and enhances the signal-to-noise ratio but typically requires additional filtering to mitigate false alarms \cite{murphy1986linear}, \cite{rey1990application}, \cite{biondi2017low}. Similarly, while the Hough transform is sensitive in detecting linear features, its high false alarm rate necessitates integration with other preprocessing techniques, such as bandpass filtering, nonlinear amplification, or digital terrain modeling, to minimize land interference and improve detection reliability \cite{srivastava2022analysis}, \cite{jiaqiu2011novel}, \cite{hong2022target}. Additionally, polarization enhancement techniques leverage polarization information from SAR imagery, enhancing wake visibility by exploiting differences in scattering properties between wakes and adjacent sea surfaces. Techniques such as the Polarization Whitening Filter and Polarization Differential Offset Filter \cite{jiang2024ship}, along with advanced methods like two-stage low-rank plus sparse decomposition (LRSD) and polarization LRSD \cite{biondi2017low}, \cite{liu2021novel}, effectively reduce clutter and enhance classification accuracy. Nevertheless, traditional linear extraction methods may face significant challenges in complex marine environments.

\begin{figure*}
  \centering 
  \includegraphics[width=\textwidth]{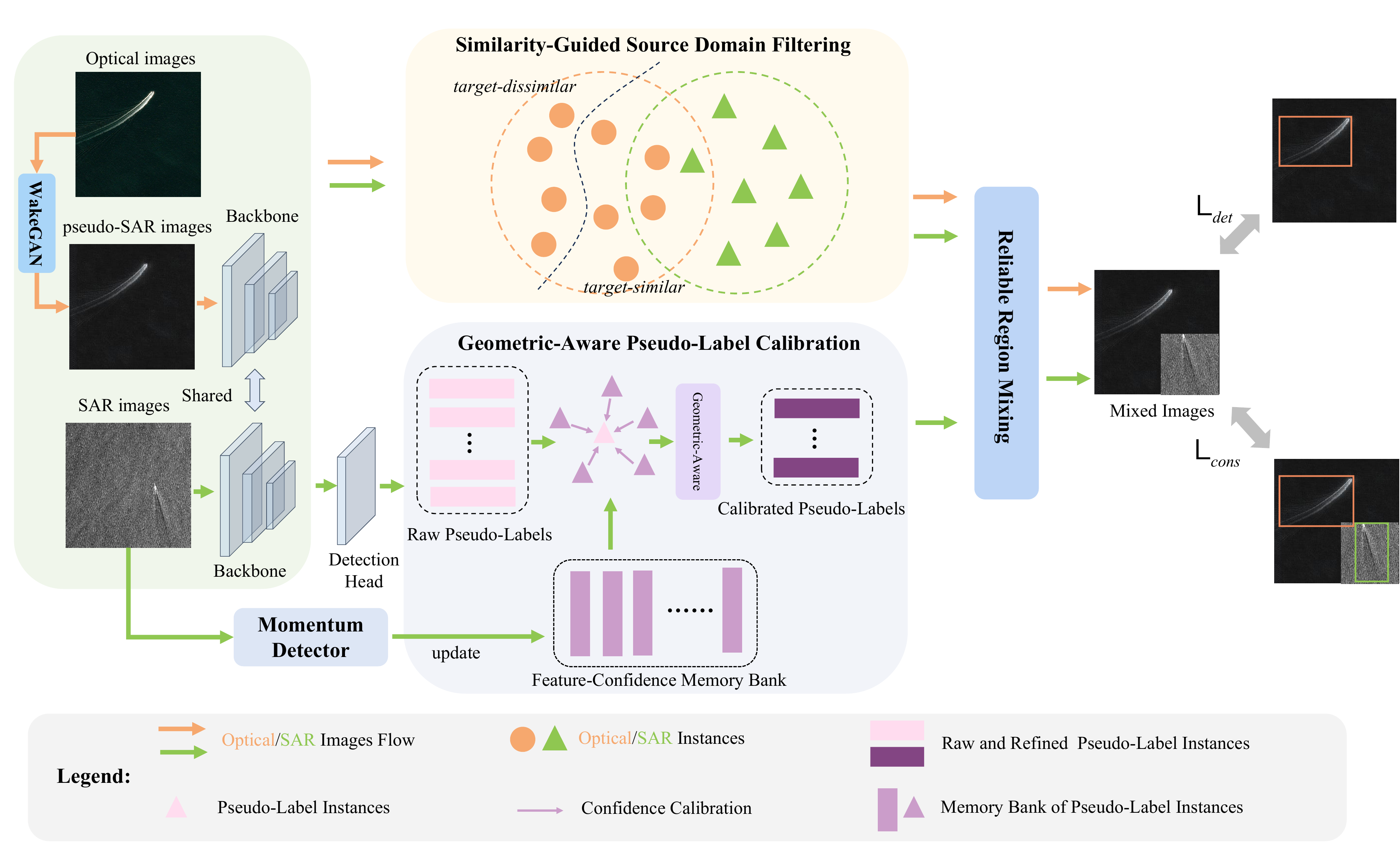} 
  \caption{Overview of the SimMemDA framework. The entire framework achieves cross-domain SAR wake detection through several steps, including wake structure-preserving domain style transfer, similarity-guided source domain filtering, memory-guided geometric-Aware pseudo-label calibration, and region-level image mixing. SimMemDA utilizes WakeGAN to convert optical images into SAR-style pseudo-images, thus reducing visual discrepancies between domains. Subsequently, source domain features similar to those of the target domain are selected through a similarity metric, thereby mitigating training noise; Calibration of pseudo-labels in the target domain is further achieved by leveraging a feature-confidence memory bank together with wake slenderness and directional consistency priors, thereby enhancing the reliability of the labels. Finally, regions of high confidence in the target domain images are selected based on confidence scores and mixed with source domain images to generate mixed images for domain adaptive training.} 
  \label{fig:model_structure} 
\end{figure*}

Recent advances in deep learning have demonstrated substantial potential for ship wake detection, particularly through convolutional neural networks and object detection frameworks. Various research efforts have successfully applied deep learning to automate and optimize wake detection. Kang et al. \cite{kang2019ship} automated ship wake detection using azimuth displacement measurements for reliable speed estimation, enhancing maritime surveillance capabilities. Ding et al. \cite{ding2023towards} optimized SAR wake detection using combined convolution modules and an angle-related loss function, achieving improved localization accuracy. Liu and Zhao\cite{liu2024kelvin} combined GoogLeNet classification with a new sub-image segmentation strategy, achieving high recall and high precision in Kelvin wake detection. Xu et al. \cite{xu2024wake2wake} introduced a novel Ship Wake Awareness module to better capture turbulent and Kelvin wake features, markedly improving detection performance in SAR imagery. Despite these advancements, deep learning approaches still face constraints due to the requirement for extensive annotated datasets. The inherent complexity and abstraction of SAR wake structures make dataset annotation tedious and labor-intensive, thus limiting the availability of large-scale annotated datasets and presenting significant challenges for supervised learning methods.

\subsection{Unsupervised domain adaptation object detection for remote sensing images}
To overcome significant differences in data distributions and imaging mechanisms between optical and SAR imagery, as well as to address annotation scarcity in SAR images, numerous cross-domain adaptation methods have emerged within remote sensing research. These methods operate at pixel, feature, and prediction levels, facilitating knowledge transfer and feature alignment through diverse strategies. Shi et al. \cite{shi2022unsupervised} introduced a progressive transfer-based unsupervised domain adaptation method, gradually reducing appearance discrepancies between optical and SAR images. He et al. \cite{he2023cross} developed a cross-modal feature transfer method, employing multi-level modality alignment and a hard-sample supervision module to effectively transfer rich RGB features to SAR images. Yuan et al. \cite{yuan2023adaptive} proposed an Adaptive Ship Detection method to progressively minimize distribution disparities across pixel, feature, and classifier levels, achieving efficient cross-domain knowledge transfer. Chen et al. \cite{xi2024cromoda} designed CroMoDa, utilizing cross-modal distribution alignment and despeckling of low-level features, significantly enhancing directional SAR ship detection accuracy. Liu et al. \cite{liu2024confidence} presented a confidence-driven region mixing strategy within a mean teacher framework, effectively mitigating detection performance degradation inherent in domain adaptation scenarios. Shi et al. \cite{shi2024unsupervised} explicitly decomposed domain-invariant and domain-specific features and employed uncertainty-guided self-training to enhance SAR object detection robustness. Yang et al. \cite{yang2024unsupervised} introduced a cross-domain feature interaction and data contribution balancing method, successfully narrowing feature gaps between optical and SAR domains. Luo et al. \cite{luo2024sar} proposed SAR-CDSS, incorporating data augmentation through image mixing and instance exchange combined with adaptive feature filtering under semi-supervised settings, enabling effective cross-domain detection. Huang et al. \cite{huang2024domain} proposed a domain adaptation framework emphasizing directional object detection, facilitating efficient knowledge transfer via pixel-level, instance-level, and multi-scale feature extraction mechanisms from optical to SAR imagery. Although these methods have effectively mitigated domain differences between optical and SAR imagery, challenges persist in pseudo-label-driven adaptation approaches. The unreliability inherent in pseudo-labels can cause error accumulation in the target domain, and significant source-target domain differences frequently introduce noise, adversely affecting detection accuracy. To address these challenges, this paper proposes a novel method integrating reliable pseudo-label filtering and effective source domain data filtering strategies. This approach aims to mitigate pseudo-label uncertainties, minimize noise introduction, significantly reduce domain discrepancies, and consequently enhance the accuracy and robustness of SAR ship wake detection.

\begin{figure*}
  \centering 
  \includegraphics[width=\textwidth]{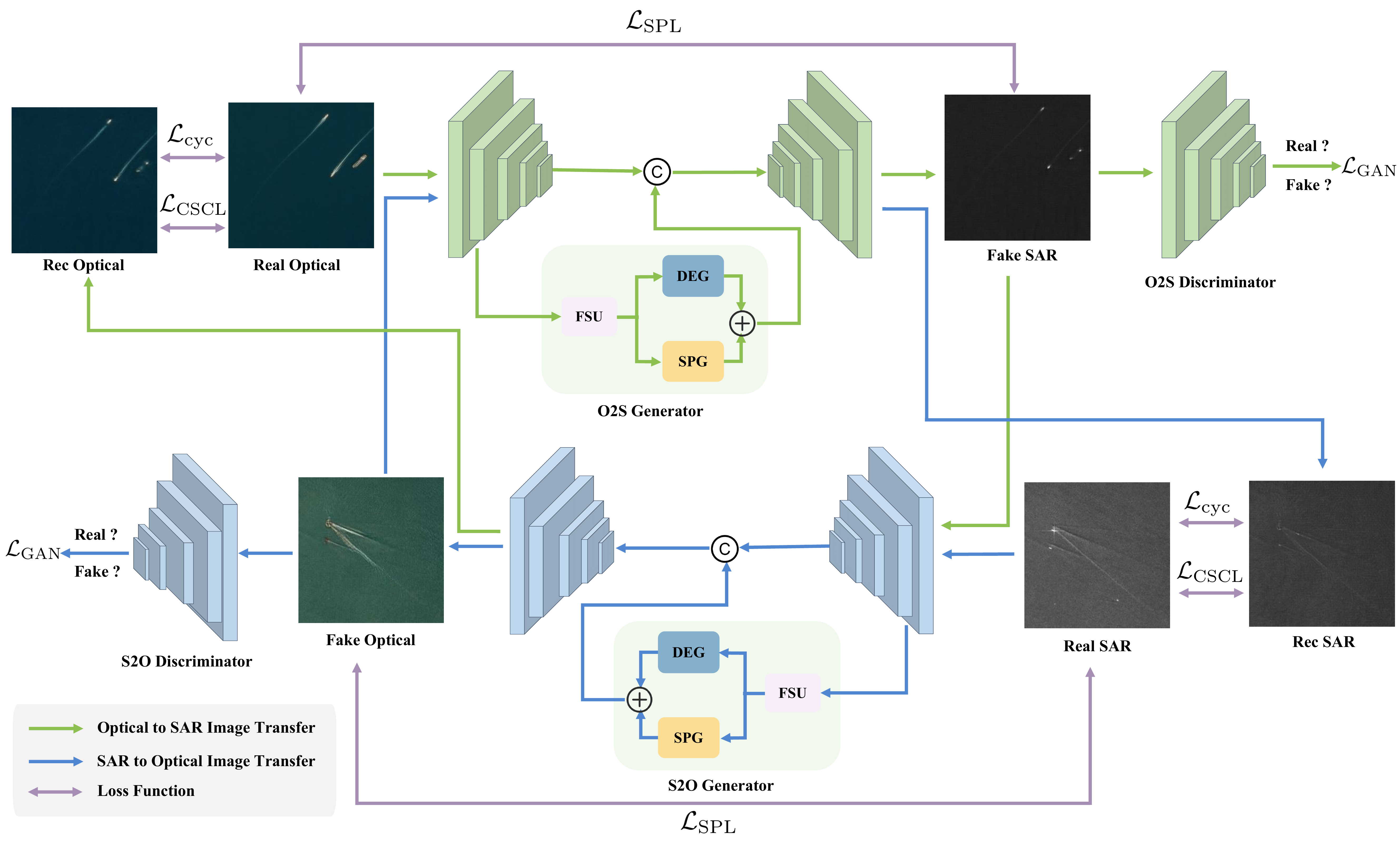} 
  
  \caption{The proposed overall architecture of WakeGAN. The model consists of two generators for Optical→SAR and SAR→Optical translation, along with their corresponding discriminators. Within each generator, a Frequency Selection Unit (FSU) decomposes features into low- and high-frequency branches, the Detail Enhancement Guide (DEG) strengthens high-frequency scattering and edge textures, and the Structure Preserving Guide (SPG) constrains low-frequency structures to maintain wake geometry. By integrating adversarial loss, cycle-consistency loss, spectral preservation loss, and cyclic spectral consistency loss, the model achieves geometrically stable transformations while generating realistic SAR wake-style textures.} 
  \label{fig:WakeGAN} 
\end{figure*}

\section{Proposed Method}\label{section_3}
Let $\mathcal{S}=\left\{\left(x_i^S, y_i^S\right)\right\}_{i=1}^{N_S}$ denote the labeled source domain, where each $x_i^S$ is an optical image and $y_i^S$ its corresponding ship wake label. Let $\mathcal{T}=\left\{x_j^T\right\}_{j=1}^{N_T}$ denote the unlabeled target domain, consisting of SAR images. The objective is to learn a detection model $F(\cdot)$ that can accurately predict ship wake locations in the target domain despite a substantial domain shift between $\mathcal{S}$ and $\mathcal{T}$.
To address the problem of unsupervised domain adaptation wake detection from optical to SAR, we propose an unsupervised domain adaptation framework for wake detection based on instance-level feature similarity filtering and feature memory guidance, as shown in Fig. \ref{fig:model_structure}. This framework fully leverages the uniqueness of wake targets in terms of geometric morphology, spectral characteristics, and linear structures. Through a multi-stage collaborative strategy, it progressively narrows the cross-domain gap and improves pseudo-label quality from the input level to the training level, thereby achieving robust detection performance. Specifically, we first design a structure-preserving style transfer model, WakeGAN, where the generation process is jointly constrained by a frequency selection unit, a detail enhancement guide, and a structure preserving guide. In addition, spectral preservation loss and cyclic spectral consistency loss are incorporated to ensure that optical wakes can be faithfully transferred into the SAR style in both geometric structures and physical textures. Furthermore, we introduce a similarity-guided source domain filtering mechanism to select source domain samples whose feature distributions are most similar to those of the target domain, thereby avoiding negative transfer. To address the pseudo-label noise problem in the unlabeled target domain, we propose a memory-guided geometric-aware pseudo-label correction strategy, which enhances pseudo-label robustness through a feature-confidence memory bank and a neighborhood fusion confidence mechanism, while leveraging wake structural priors for geometric consistency correction. Subsequently, we employ a region selection and sample mixing strategy to achieve mixed supervision between annotated source domain samples and refined pseudo-labels from the target domain, thereby enhancing cross-domain feature adaptation capability. Finally, under the adaptive detector training paradigm, we combine supervised detection loss with self-supervised consistency loss to progressively improve the generalization performance of the detector in the target domain.

\subsection{Structure-Preserving Domain Style Transfer}\label{section_31}
In the task of domain adaptive wake detection between optical and SAR modalities, optical sensors and SAR sensors exhibit fundamental differences in imaging mechanisms, signal statistical properties, and noise patterns. Wakes in optical images rely on visible light reflection, presenting high-resolution and color rich appearance features, while wakes in SAR images are formed by radar microwave echoes, characterized by speckle-noise dominance, pronounced anisotropic textures, and prominent strong scattering points. Such inter domain differences manifest as significant distribution shifts in the feature space, directly causing detectors trained in the optical domain to suffer severe performance degradation in the SAR domain. For wakes as a phenomenon, in the optical domain they appear as long-scale grayscale gradients and water surface disturbances, whereas in the SAR domain they manifest as high-low contrast stripes and complex scattering textures generated by the interaction of radar waves with sea surface roughness. Under the condition of significant discrepancies between source and target domains, merely relying on feature space distribution alignment is insufficient to eliminate the pronounced pixel distribution shift caused by different imaging mechanisms, particularly in the cross-modal scenario from optical to SAR. Therefore, at the pre-training stage of detection, an input-level domain alignment strategy is introduced. By performing optical to SAR style transfer to generate synthetic wake images with SAR style while preserving the geometric structures of the optical images, the statistical differences between source and target domains can be effectively reduced. This not only lowers the learning difficulty of the feature extractor but also preserves the usability of the original labels, thereby providing task-friendly training samples for the subsequent detection model.

 However, compared with general image-to-image translation tasks, wake-specific style transfer faces distinct challenges \cite{chen2022pixel},\cite{pu2023ship}. In terms of geometric structure preservation, the morphology of wakes (including scale, orientation, and position) directly determines the validity of annotations. Any geometric deformation, scale drift, or positional shift introduced during the generation process would break the semantic alignment between original labels and synthetic images, thereby diminishing the value of samples. Regarding frequency-domain consistency, SAR wake signals simultaneously contain low-frequency global modulation and high-frequency local disturbances, while the statistical distributions of these frequency components differ significantly across domains. To address the above issues, we design a wake domain adaptive structure-preserving style transfer model, WakeGAN, which is built upon CycleGAN \cite{zhu2017unpaired}. The framework embeds three types of guiding components into the generator and introduces spectral consistency losses to enforce dual constraints on both geometry and physical textures, as shown in Fig. \ref{fig:WakeGAN}. The model consists of generators $G_{\mathcal{S}}:\mathcal{S} \rightarrow \mathcal{T}$ and $G_{\mathcal{T}}:\mathcal{T} \rightarrow \mathcal{S}$, as well as discriminators $D_{\mathcal{T}}$ and $D_{\mathcal{S}}$. Within the generator, a Frequency Selection Unit (FSU) is embedded to adaptively decompose shallow features into low- and high-frequency branches; a Detail Enhancement Guide (DEG) is applied to the high-frequency branch to reinforce local scattering and edge details; and a Structure Preserving Guide (SPG) is applied to the low-frequency branch to stabilize large-scale wake morphology. Meanwhile, to further constrain the cross-domain spectral distribution consistency, WakeGAN introduces a Spectral Preservation Loss (SPL) and a Cyclic Spectral Consistency Loss (CSCL), which jointly restrict spectral drift under cyclic constraints, thereby strictly preserving geometric structures while restoring physical textures in the target domain.

\begin{figure*}
  \centering
  \includegraphics[width=0.8\textwidth]{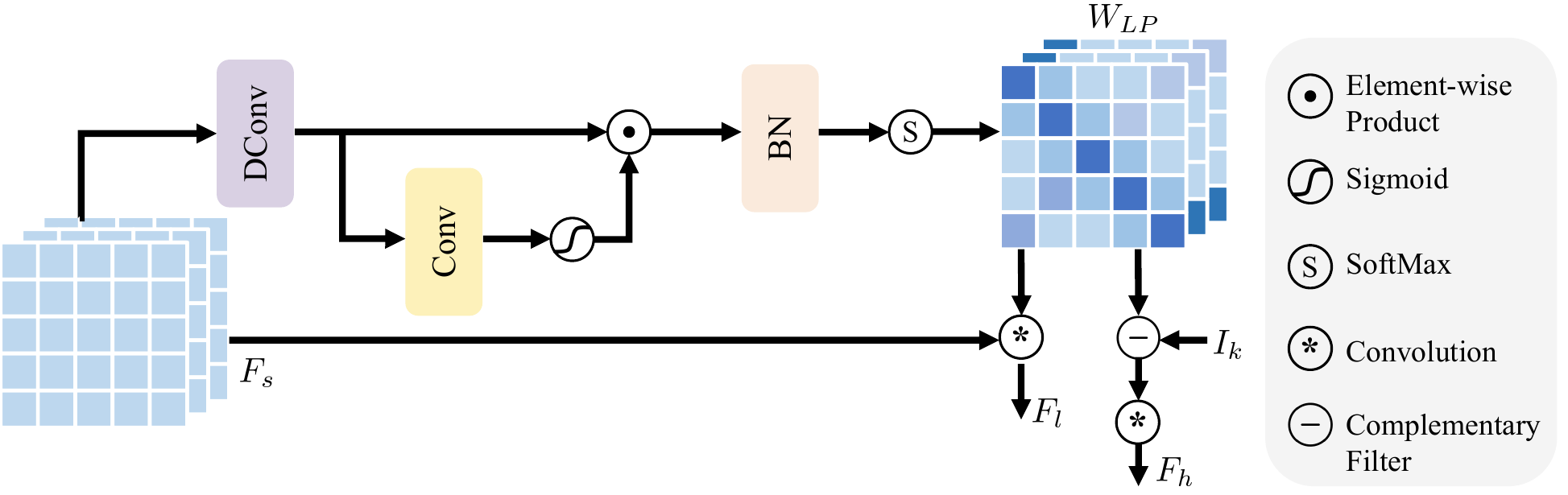}
  \caption{Structure diagram of the FSU.}
  \label{fig:FSU}
\end{figure*}

\begin{figure*}
  \centering
  \includegraphics[width=0.8\textwidth]{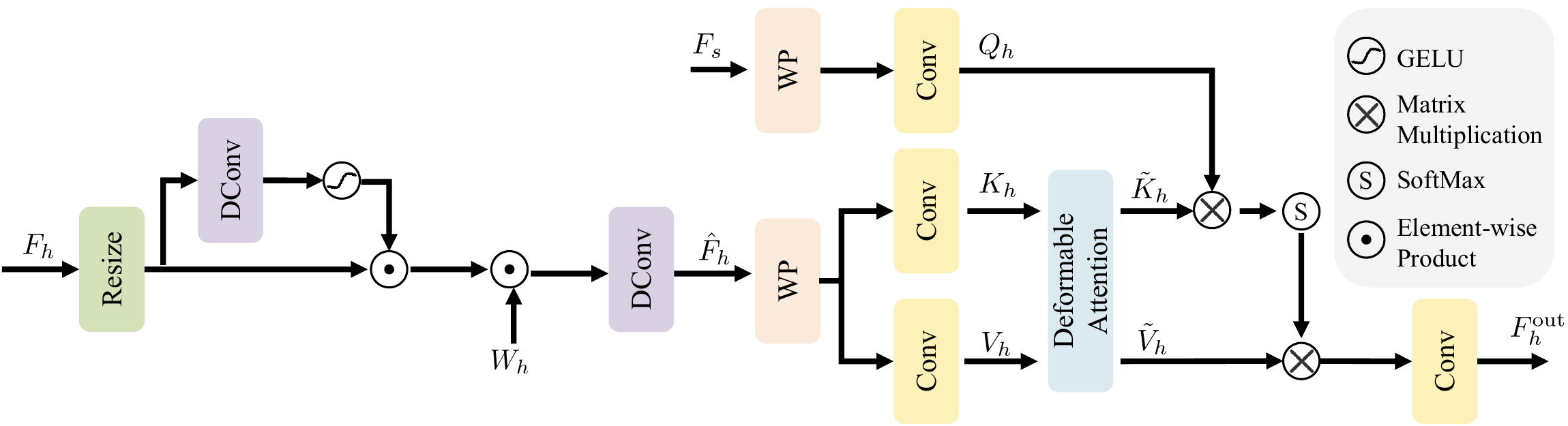}
  \caption{Structure diagram of the DEG.}
  \label{fig:DEG}
\end{figure*}

Specifically, let $x^S$ denote an image from the source (optical) domain $\mathcal{S}$, and $x^T$ denote an image from the target (SAR) domain $\mathcal{T}$. The shallow features of the generator backbone $G_{\mathcal{S}}$ are represented as $F_s \in \mathbb{R}^{H \times W \times C}$. 

\subsubsection{Frequency Selection Unit}
The frequency selection unit (FSU) adaptively decomposes the shallow input features $F_s$ into low-frequency components $F_l$ and high-frequency components $F_h$, while preserving the global geometric consistency of the features. This ensures that the low-frequency branch focuses on smooth structural modeling, whereas the high-frequency branch emphasizes texture and edge enhancement, as shown in Fig. \ref{fig:FSU}. Specifically, the feature channels are divided into $g$ groups, each containing $C_g=C / g$ channels. For each group of features $F_s^{(i)}$, spatial information is first extracted via depthwise separable convolution, followed by nonlinear activation to obtain the weight prototype $\hat{Z}$:
\begin{equation}
\hat{Z}^{(i)}=\phi\left(\operatorname{DConv}_{3 \times 3}^{(i)}\left(F_s^{(i)}\right)\right)
\end{equation}
where $\phi$ denotes the Sigmoid activation function. Subsequently, to further enhance discriminability, it is multiplied channel-wise with the result mapped by a $1 \times 1$ convolution
\begin{equation}
\tilde{Z}^{(i)}=\hat{Z}^{(i)} \odot \phi\left(\operatorname{Conv}_{1 \times 1}^{(i)}(\hat{Z}^{(i)})\right)
\end{equation}

Based on this, $\tilde{Z}^{(i)}$ is batch-normalized and then passed through a Softmax to obtain the normalized low-pass filtering kernel
\begin{equation}
W_{\mathrm{LP}}^{(i)}=\operatorname{Softmax}(\mathrm{BN}(\tilde{Z}^{(i)}))
\end{equation}

Applying this low-pass kernel to the input features yields the low-frequency components
\begin{equation}
F_l=\left\{W_{\mathrm{LP}}^{(i)} * F_s^{(i)}\right\}_{i=1}^g
\end{equation}
where $*$ denotes the convolution operation. To obtain the complementary high-frequency components, a delta kernel is used to construct a high-pass filter, which is then applied to the input features
\begin{equation}
F_h=\left\{\left(I_k^{(i)}-W_{\mathrm{LP}}^{(i)}\right) * F_s^{(i)}\right\}_{i=1}^g
\end{equation}
where $I_k$ is the delta kernel, ensuring that the high-pass and low-pass kernels are complementary in the spatial domain.

\begin{figure*}
  \centering
  \includegraphics[width=\textwidth]{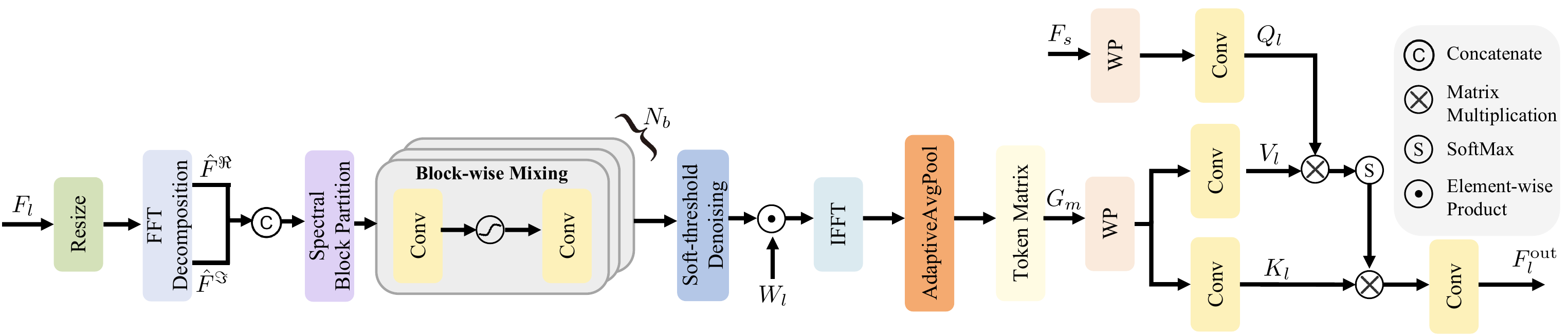}
  \caption{Structure diagram of the SPG.}
  \label{fig:SPG}
\end{figure*}

\subsubsection{Detail Enhancement Guide}

In the cross-domain wake detection task, high-frequency detailed features directly determine edge localization, texture distribution, and the geometric representation of small targets. In contrast, low-frequency components govern the consistency of the wake with the sea surface background, energy distribution, and overall perceptual style. If high-frequency information is missing, the detector's localization accuracy and class discrimination ability will significantly degrade. During the style transfer process, since low-frequency components have larger magnitudes, they often overshadow the contribution of high-frequency components in similarity computation, making it difficult for attention to effectively focus on structurally salient regions. To address this, we propose the Detail Enhancement Guide (DEG), which enhances the capability of attention in detail saliency modeling and geometric deformation adaptability through a two-stage guidance mechanism, as shown in Fig. \ref{fig:DEG}. Specifically, DEG first introduces a High-Frequency Guided Modulation (HFGM) mechanism, which performs joint spatial-channel weighting on the high-frequency branch features, thereby increasing the weight of structurally salient regions in similarity computation and suppressing weakly relevant or noisy regions. The high-frequency features $F_h$ extracted from the FSU are resized to match the spatial resolution of the attention window $(\hat{H}, \hat{W})$, resulting in a scale-consistent feature representation $\tilde{F}_{h}=\operatorname{Resize}(F_{h} ; \hat{H}, \hat{W})$. Subsequently, we introduce a lightweight spatial-channel decoupled convolution operator $\mathcal{G}(\cdot)$ to extract structural responses from the features
\begin{equation}
A=\sigma\left(\mathcal{G}\left(\tilde{F}_{h}\right)\right)
\end{equation}
where $\sigma(\cdot)$ denotes the GELU activation function, and $\mathcal{G}(\cdot)$ consists of two sets of depthwise separable convolutions in orthogonal directions. To further incorporate prior structural preference at both spatial and channel levels, we define a learnable template $W_h \in \mathbb{R}^{C \times \hat{H} \times \hat{W}}$, which adaptively learns the importance distribution of different positions and channels during training. The output of the high-frequency guided modulation is given by
\begin{equation}
\hat{F}_h =\operatorname{DConv}_{3 \times 3}\left(\left(A \odot \tilde{F}_{h}\right) \odot W_h\right)
\end{equation}

To maximize the utilization of structurally modulated high-frequency information and suppress alignment errors caused by subtle geometric deformations in cross-domain scenarios, we incorporate deformable sampling into window attention and propose a High-Frequency Guided Deformable Window Attention mechanism. During the sampling stage of key and value features, continuous two-dimensional offsets guided by the query are introduced, enabling adaptive resampling of key and value feature positions beyond the fixed regular grid, thereby preventing attention weights from spreading to non-corresponding positions. Specifically, the query matrix $Q_h$ is derived from the input shallow features $F_s$ to preserve the original geometric semantics, $Q_h=W_p^Q \cdot WP(F_s)$; whereas the key $K_h$ and value $V_h$ are generated based on the enhanced high-frequency guided features, $K_h=W_p^K \cdot WP(\hat{F}_h)$, $V_h=W_p^V \cdot WP(\hat{F}_h)$, where $W_p^{(\cdot)}$ denotes the learnable projection matrix and $WP(\cdot)$ denotes the window partition strategy\cite{liu2021swin}.

Next, during the feature sampling stage of $K_h$ and $V_h$, we introduce two-dimensional offset prediction guided by $Q_h$. A predictor $\Psi(\cdot)$, composed of a depthwise separable convolution and $1 \times 1$ convolution, is employed to generate the corresponding offset field:
\begin{equation}
\Delta p =\tanh \left(\Psi\left(Q_h\right)\right) \cdot s
\end{equation}
where $s$ is a normalization scaling factor, ensuring that the offsets are controllable within the range $[-1,1]$. Let the regularized base grid be $G_0 \in[-1,1]^{M \times M \times 2}$, then the corrected sampling positions are $\hat{p}= G_0 +\Delta p$. For any target position $\hat{p}^{(u, v)}=\left(\hat{p}_{u, v}^x, \hat{p}_{u, v}^y\right)$, we employ a bilinear sampling operator
\begin{equation}
\begin{split}
\mathcal{S}\left(X, \hat{p}^{(u, v)}\right) 
&= \sum_{\left(r_x, r_y\right) \in \mathcal{N}(\hat{p}^{(u, v)})} 
    g\left(\hat{p}_{u, v}^x, r_x\right) \\
&\quad \cdot g\left(\hat{p}_{u, v}^y, r_y\right) 
    \cdot X\left[:, r_y, r_x\right]
\end{split}
\end{equation}
where $\mathcal{N}\left(\hat{p}^{(u, v)}\right)$ denotes the four nearest integer grid points around the target position $\hat{p}^{(u, v)}$, $g(a, b)=\max (0,1-|a-b|)$ is the bilinear kernel function; $r_x$ and $r_y$ represent the integer grid indices in the horizontal and vertical directions within the sampling neighborhood.

By applying the above sampling operator to $K_h$ and $V_h$, the offset-integrated key $\tilde{K}_h$ and value $\tilde{V}_h$ can be obtained $\tilde{K}_h=\mathcal{S}\left(K_h, \hat{p}\right)$, $\tilde{V}_h=\mathcal{S}\left(V_h, \hat{p}\right)$. Subsequently, the attention is computed in the standard scaled dot-product form
\begin{equation}
O_h= \tilde{V}_h \cdot \operatorname{Softmax}\left(\frac{Q_h \cdot \tilde{K}_h^{\top}}{\sqrt{d}}\right) 
\end{equation}
where $d$ denotes the dimension of a single head. 

The outputs of multiple heads are concatenated along the channel dimension and then projected back to the original channel dimension through a $1 \times 1$ convolution. Finally, the local results are restored to the global spatial layout via the window reverse transformation, yielding the final feature $F_{h }^{\text {out }}$.

\subsubsection{Structure Preserving Guide}

To achieve reliable style transfer from optical wakes to SAR wakes, we propose a Structure Preserving Guide (SPG), which focuses on constraining the processing in the low-frequency stage, as shown in Fig. \ref{fig:SPG}. High-frequency components determine the fidelity of wake details and the clarity of geometric edges, while low-frequency components shape the consistency between the wake and the sea surface background, illumination and scattering patterns, as well as the overall perceptual style. If the low-frequency processing is inappropriate, even with accurate high-frequency transfer, perceptual disharmony may arise due to background misalignment or mismatched cross-domain energy distribution, thereby weakening the detector's effective utilization of transferred samples. To address this, the SPG performs low-frequency structural modeling in the frequency domain through block-wise mixing and learnable soft-threshold denoising, and incorporates a multi-token cross-attention mechanism to adaptively select the optimal low-frequency guidance across different spatial regions. This design not only ensures that the statistical differences between wakes and background remain within a controllable range, but also effectively preserves geometric layouts and long-range structures, thereby avoiding common issues in traditional low-frequency fusion such as wake discontinuities or background flooding. Specifically, we extract the low-frequency component features $F_l$, resize them to obtain $\tilde{F}_{l} \in \mathbb{R}^{\hat{H} \times \hat{W} \times C}$, then apply a two-dimensional Fourier transform $\hat{F}_{l}=\mathcal{F}(\tilde{F}_{l})$, and concatenate the real and imaginary parts of the complex spectrum along the channel dimension into $2C$ channels, yielding $\bar{F_l} \in \mathbb{R}^{B \times 2 C \times H}$.

Considering that low-frequency energy often exhibits narrowband clustering and directional anisotropy, we divide $\bar{F_l}$ along the spectral width into $N_b$ non-overlapping blocks $\left\{\mathcal{B}_i\right\}_{i=1}^{N_b}$, where each block has a width $w_i$ satisfying $\sum_i w_i=W_f$. For each block $\bar{F_l}_i \in \mathbb{R}^{B \times 2 C \times H}$, block-wise mixing is performed to encode cross-channel and cross-band coupling relationships,
\begin{equation}
Z_i=W_{2i}\left(\phi\left(~W _{1i}\left(\bar{F_l}_i\right)\right)\right),
\end{equation}
where $\phi$ denotes the GELU activation function. All blocks are concatenated along the spectral width direction to obtain $\hat{Z}=\operatorname{concat}\left(Z_1, \ldots, Z_{N_b}\right)$. To suppress noise amplification at low-energy frequency points and enhance training stability, we introduce a learnable soft-threshold $\tau \in \mathbb{R}^{1 \times 2 C \times 1 \times 1}$ and apply channel-wise soft-thresholding on $\hat{Z}$:
\begin{equation}
\mathcal{S}(\hat{Z} ; \tau)=\operatorname{sign}(\hat{Z}) \odot \max (|\hat{Z}|-\tau, 0)
\end{equation}

The $\mathcal{S}(\hat{Z} ; \tau)$ is then decomposed into the real part $\hat{Z}^{\Re}$ and the imaginary part $\hat{Z}^{\Im}$. After recombination in the complex domain with a learnable template $W_l \in \mathbb{R}^{C \times H_f \times W_f}$, the inverse Fourier transform is applied to return to the spatial domain
\begin{equation}
\hat{F_{l}}=\mathcal{F}^{-1}(\left(\hat{Z}^{\Re}+i \hat{Z}^{\Im}\right) \odot W_l)
\end{equation}

Next, we divide $\hat{F_{l}}$ into a $g_h \times g_w$ grid of features using adaptive average pooling, obtaining $G=\operatorname{AdaptiveAvgPool}_{\left(g_h, g_w\right)}\left(\hat{F_{l}}\right) \in \mathbb{R}^{B \times C \times g_h \times g_w}$. Each spatial cell is flattened into a $C$-dimensional token, forming a compact low-frequency token matrix $G_M \in \mathbb{R}^{B \times M \times C}$, where $M=g_h g_w$. These tokens simultaneously possess localization and frequency-aware properties, facilitating the mapping between spatial positions and their corresponding low-frequency patterns. Subsequently, we apply a $1 \times 1$ convolution to the shallow features $F_s$ and flatten them as the query $Q_l=W_q^Q \cdot WP(F_{s})$, while the low-frequency tokens generate the corresponding key $K_l = W_q^K \cdot G_M$ and value $V_l = W_q^V \cdot G_M$. Then, the multi-head dot-product attention mechanism is applied
\begin{equation}
O_l=V_l \cdot \operatorname{Softmax}\left(\frac{Q_l \cdot K_l^{\top}}{\sqrt{d}}\right) 
\end{equation}
which enables the model to integrate global contextual dependencies across low-frequency patterns. Finally, after dimension rearrangement to restore the spatial feature map, a linear projection layer maps the features back to the backbone feature space, producing the low-frequency output feature $F_{l }^{\text {out }}$.

\subsubsection{Loss Function}
To further enhance frequency-domain consistency and prevent drift in the global low-frequency structure as well as distortion of high-frequency directional features, we design the Spectral Preservation Loss (SPL), which explicitly constrains the spectral stability between input and output during the generation process
\begin{equation}
\begin{aligned}
\mathcal{L}_{\mathrm{SPL}}(G_{\mathcal{S}},x^S) 
&= \mathbb{E}_{x^S \sim \mathcal{S}}\Big[ 
    \left\|p_L\left(x^S\right) - p_L\left(G_{\mathcal{S}}\left(x^S\right)\right)\right\|_2^2 \\
&\quad + \lambda_H \, d_H\left(p_H\left(x^S\right), p_H\left(G_{\mathcal{S}}\left(x^S\right)\right)\right) 
\Big]
\end{aligned}
\end{equation}
where $p_L(\cdot)$ and $p_H(\cdot)$ denote the low-frequency and high-frequency components extracted via polar coordinate histogram binning \cite{li2022robust},the low-frequency term adopts the $L_2$ distance to measure global structural differences, while the high-frequency term employs directional cosine distance $d_H(\cdot,\cdot)$ to evaluate consistency in directional distribution; $\lambda_H$ is the weighting factor for the high-frequency term.

Traditional cycle consistency only constrains reconstruction errors in the pixel domain, making it difficult to ensure the stability of frequency-domain features in the forward-backward mapping, particularly with respect to the alignment of high-frequency patterns and low-frequency structures. Therefore, we further design the Cyclic Spectral Consistency Loss (CSCL), which introduces spectral alignment constraints into the cyclic path to enforce stability and reversibility of frequency-domain signatures during the round-trip mapping, thereby suppressing spectral drift, directional inversion, and the accumulation of high-frequency noise
\begin{equation}
\begin{aligned}
\mathcal{L}_{\mathrm{CSCL}}(G_{\mathcal{S}}, G_{\mathcal{T}}) 
&= \mathbb{E}_{x^S \sim \mathcal{S}}\Big[
    \left\|p_L\left(x^S\right) - p_L\left(\bar{x}^S\right)\right\|_2^2 \\
&\quad + \lambda_H \, d_H\left(p_H\left(x^S\right), p_H\left(\bar{x}^S\right)\right)
\Big] \\
&\quad + \mathbb{E}_{x^T \sim \mathcal{T}}\Big[
    \left\|p_L\left(x^T\right) - p_L\left(\bar{x}^T\right)\right\|_2^2 \\
&\quad + \lambda_H \, d_H\left(p_H\left(x^T\right), p_H\left(\bar{x}^T\right)\right)
\Big]
\end{aligned}
\end{equation}
where $\bar{x}^S = G_{\mathcal{T}}(G_{\mathcal{S}}(x^S))$, and $\bar{x}^T = G_{\mathcal{S}}(G_{\mathcal{T}}(x^T))$.

Finally, the loss function also includes adversarial loss and cycle consistency loss \cite{zhu2017unpaired}, yielding the overall objective function of WakeGAN

\begin{equation}
    \label{eq:cyclegan_loss}
    \begin{aligned}
    \mathcal{L}_{\mathrm{WakeGAN}}
      &= \mathcal{L}_{\mathrm{SPL}}(G_{\mathcal{S}},x^S) + \mathcal{L}_{\mathrm{SPL}}(G_{\mathcal{T}},x^T) \\
      &\quad +\mathcal{L}_\mathrm{GAN}\bigl(G_{\mathcal{S}}, D_{\mathcal{T}}\bigr)
         + \mathcal{L}_\mathrm{GAN}\bigl(G_{\mathcal{T}}, D_{\mathcal{S}}\bigr) \\
      &\quad + \mathcal{L}_{\mathrm{CSCL}}(G_{\mathcal{S}}, G_{\mathcal{T}})  +
      \lambda_{\text{CYC}}\,
        \mathcal{L}_{\mathrm{CYC}}\bigl(G_{\mathcal{S}}, G_{\mathcal{T}}\bigr)
    \end{aligned}
    \end{equation}

After training the WakeGAN, we use the learned generator $G_{\mathcal{S}}$ to transform all source domain images into SAR-style images. Given a source optical image $x^S \in \mathcal{S}$ with an annotated ship wake, we obtain a pseudo-SAR image $\tilde{x}^T = G_{\mathcal{S}}(x^S)$ that exhibits the visual characteristics of SAR. These pseudo-SAR images form an intermediate domain that is stylistically closer to real SAR data while preserving the content (locations and shapes of wakes) from the optical domain. In this way, we are able to reduce the differences in appearance between optical images and SAR images.

\subsection{Similarity-Guided Source Domain Filtering}\label{section_32}
In cross-domain wake detection, since some samples in the source domain often exhibit significant distribution differences compared to the target domain, directly using them for adaptive training may introduce a large amount of noise, leading to cumulative errors and negative transfer. Although style transfer makes the two domains visually closer, mere appearance transformation cannot fully resolve the distribution differences between the source and target domain data. To address this, we filter the source domain data, retaining only samples with high similarity to the target domain to ensure that the model learns relevant information. We first use the backbone network $\mathcal{F}$ of a detector to extract feature representations for each instance. For any input image $x_i^s \in \mathcal{S}$ and $x_j^t \in \mathcal{T}$, their feature maps are denoted as $v_i^s=\mathcal{F}\left(x_i^s\right), \quad v_j^t=\mathcal{F}\left(x_j^t\right)$. For the $j$-th instance in the target domain, its bounding box in the image is denoted as $c_j^t$. We define a cropping function $H(\cdot)$ to extract the corresponding instance features from $v_j^t$ based on $c_j^t$. The extracted features are then processed with average pooling to obtain the instance-level embedding of the target domain $\phi_j^t$:
\begin{equation}
  \phi_j^t=\operatorname{AvgPooling}\left(H\left(v_j^t, c_j^t\right)\right),
  \end{equation}
Similarly, the instance representation $\phi_i^s$ in the source domain $\mathcal{S}$ can also be obtained using the same method with $c_i^s$ and $v_i^s$.

Then, we aim to estimate the potential distribution of target instances using the embeddings of all instances in the target domain $\Phi_t=\left\{\phi_j^t\right\}$. Specifically, we introduce a parameterized distribution model $\Theta\left(\theta_t\right)$, where $\theta_t$ represents the learnable distribution parameters. To enable $\Theta\left(\theta_t\right)$ to better describe the target domain data, we solve for $\theta_t$ by maximizing the log-likelihood of target instances under this distribution:
\begin{equation}
  \theta_t=\arg \max _\theta \sum_{\phi^t \in \Phi_t} \log \left(\Theta\left(\phi_j^t \mid \theta\right)\right),
  \end{equation}
where $\Theta\left(\phi_j^t \mid \theta_t\right)$ represents the distribution corresponding to the parameter $\theta_t$, and $\phi_j^t$ denotes probability.

Due to the differences between domains, if we can filter out the portion of source domain data that is similar to the target domain distribution, these samples can be better utilized in subsequent domain adaptive training. To achieve this, we introduce a similarity metric based on the target distribution to select source domain data that are most similar to the target domain and refer to them as the \emph{target-similar} source domain set. The remaining samples are considered as the \emph{target-dissimilar} source domain set. Specifically, for any source domain instance $\phi_i^s$, we take the reciprocal of the probability of this instance occurring under the target distribution as a distance metric:
\begin{equation}
  \label{equ_distance}
  d_i^s=\frac{1}{\Theta\left(\phi_i^s \mid \theta_t\right)},
\end{equation}
Then, we sort $d_i^s$ in ascending order and select the top $|\Phi_s^{\prime}|= \eta \cdot\left|\Phi_s\right|$ source instances based on a proportion $\eta$. The parameter $\eta$ controls the error introduced by $\Phi_s^{\prime}$, which constitutes the set of source instances most likely to be similar to the target domain distribution.

To estimate $\Theta(\theta_t)$, we consider several discrepancy estimation functions, including the $L_2$ distance with average pooling, $K$-means and the Gaussian Mixture Model (GMM).

\textbf{\emph{$L_2$ distance with average pooling}}. In the simplest case, the mean vector of all instance embeddings in the target domain can be regarded as the prototype parameter $\theta_t$:
\begin{equation}
  \theta_t=\frac{1}{\left|\Phi_t\right|} \sum_{j=1}^{\left|\Phi_t\right|} \phi_j^t,
  \end{equation}
Then, for each source domain instance $\phi_i^s \in \Phi_s$, its $d_i^s$ is computed as follows:
\begin{equation}
  d_i^s=\left\|\phi_i^s-\theta_t\right\|^2.
  \end{equation}

  \textbf{\emph{$K$-means}}. We use $K$-means to cluster the target domain instances $\Phi_t$, obtaining $K$ cluster centers $\theta_t=\left\{\theta_1^t, \theta_2^t, \ldots, \theta_K^t\right\}$. For any source domain instance $\phi_i^s$, we find the nearest cluster center among them and compute the Euclidean distance:
\begin{equation}
  d_i^s=\min _{1 \leq n \leq K}\left\|\phi_i^s-\theta_n^t\right\|^2.
  \end{equation}
 
\textbf{\emph{Gaussian Mixture Model (GMM)}}. Assuming that the target-domain distribution can be represented as a weighted mixture of multiple Gaussian components, we employ a Gaussian Mixture Model defined by:
\begin{equation}
  \Theta\left(\theta_t\right)=\sum_{m=1}^M \alpha_m \mathcal{N}\left(\mu_m, \Sigma_m\right),
\end{equation}
where $\alpha_m$ represents the mixing weight of the $m$-th Gaussian component with $\sum_{m=1}^M \alpha_m=1$, and $\mu_m$ and $\Sigma_m$ denote the mean vector and covariance matrix of the $m$-th component, respectively. The parameter set $\theta_t=\left\{\alpha_m, \mu_m, \Sigma_m\right\}_{m=1}^M$ is estimated using the Expectation-Maximization (EM) algorithm \cite{dempster1977maximum}. Accordingly, for each source domain instance embedding $\phi_i^s$, the discrepancy measure is defined as:
\begin{equation}
  d_i^s=\frac{1}{\sum_{m=1}^M \alpha_m \mathcal{N}\left(\phi_i^s ; \mu_m, \Sigma_m\right)},
  \end{equation}
where $\mathcal{N}\left(\cdot ; \mu_m, \Sigma_m\right)$ denotes the Gaussian density function.

\subsection{Memory-Guided Geometric-Aware Pseudo-Label Calibration}\label{section_33}

Based on the trained detector, pseudo-labels are generated for unlabeled data in the target domain. However, the pseudo-labels themselves may suffer from a certain degree of unreliability. In particular, during the early adaptation stage, factors such as background noise and feature ambiguity can negatively affect pseudo-labels, thereby introducing noisy supervisory signals and undermining the stability of subsequent training. To address this, we propose a memory-guided geometric-aware pseudo-label calibration strategy. This mestrategythod leverages the momentum model to generate candidate instances and mitigates the uncertainty of single point predictions through neighborhood confidence mixing. Furthermore, it incorporates a geometry-aware factor to capture the structural prior of wake targets, thereby refining and calibrating the original confidence scores. On this basis, a global-instance adaptive thresholding mechanism is introduced to dynamically adjust the filtering criteria, which significantly enhances the reliability and robustness of pseudo-labels, providing higher-quality supervisory signals for cross-domain wake detection.

\subsubsection{Feature-Confidence Memory Bank}
 In order to leverage neighbor samples that are similar to the target samples, we construct a Feature-Confidence Memory Bank of length $n$, used to store the feature embeddings and predicted confidence scores of detection boxes in the target domain. During the $k$-th round of pseudo-label generation, the memory bank is denoted as $\left\{\mathcal{M}_i^k\right\}_{i=1}^{n_t}$. The Feature-Confidence Memory Bank $\left\{\mathcal{M}_i^k\right\}_{i=1}^{n_t}$ is obtained by integrating the current pseudo-label set $\left\{\mathcal{\hat{L}}_i^k\right\}_{i=1}^{n_l}$ with the historical memory bank $\left\{\mathcal{M}_i^{k-1}\right\}_{i=1}^{n_t}$. Meanwhile, for the predicted target instances output by the detector $\left\{\mathcal{B}_i^k\right\}_{i=1}^{n_b}$, to achieve high-quality and accurate target instance predictions, we draw inspiration from MoCo \cite{he2020momentum} and introduce a momentum model $\mathcal{G}^{\prime}$ running concurrently with the detection model $\mathcal{G}$. Under this framework, parameters $\theta_t^{\prime}$ of the momentum model gradually approach the parameters $\theta_t$ of the main model at a slower pace: 
 \begin{equation}
  \theta_t^{\prime} \leftarrow m \theta_t^{\prime}+(1-m) \theta_t,
  \end{equation}
 where $m \in[0,1)$ is the momentum coefficient. The momentum model $\mathcal{G}^{\prime}$ performs inference on target domain images after each iteration and outputs the predicted detection boxes $\left\{\mathcal{B}_i^k\right\}_{i=1}^{n_b}$. Upon obtaining prediction results $\left\{B_i^k\right\}$, to prevent incorrect pseudo-label assignments caused by uncertainty, we design an adaptive threshold $\tau_{k}$ for pre-screening each target instance $b \in \left\{ \mathcal{B}_i^k\right\}$ along with its corresponding predicted confidence score $c$: 
 \begin{equation}
  \left\{\hat{\mathcal{L}}_i^k\right\}=\{b \mid c \geq \tau_{k}\},
\end{equation}
That is, the pseudo-label of the target instance $b$ is retained only if its confidence is no less than the threshold $\tau_{k}$, thereby effectively reducing pseudo-label noise and improving overall quality.

In addition, through the integration of the memory bank, the pseudo-labels generated in the $k$-th epoch, denoted as $\left\{\mathcal{\hat{L}}_i^k\right\}_{i=1}^{n_l}$,  are combined with the historical memory bank $\left\{\mathcal{M}_i^{k-1}\right\}_{i=1}^{n_t}$ to update and obtain the new memory bank $\left\{\mathcal{M}_i^k\right\}_{i=1}^{n_tl}$ for subsequent model training. After completing the $k$-th epoch of pseudo-label generation, each target instance contains two attributes, namely the feature embedding $f$ and the confidence score $c$. Specifically, assuming the pseudo-label set $\left\{\mathcal{\hat{L}}_i^k\right\}$ includes $n_l$ target instances, it can be represented as $\left\{\mathcal{\hat{L}}_i^k\right\}=\left\{\left(f_{i,j}^k, c_{i,j}^k\right)\right\}_{j=1}^{n_l}$. Meanwhile, the historical memory bank $\left\{\mathcal{M}_i^{k-1}\right\}$ contains $n_t$ target instances, expressed as $\left\{\mathcal{M}_i^{k-1}\right\} =\left\{\left(f_{i,j}^{k-1}, c_{i,j}^{k-1}\right)\right\}_{j=1}^{n_t}$. To ensure the memory bank promptly reflects the latest detection results, we design the following update strategy: for any new target instance $\left(f_{i,j}^k, c_{i,j}^k\right)$ from $\left\{\mathcal{\hat{L}}_i^k\right\}$, we first perform retrieval matching according to the sample of the target instance in the historical memory bank $\left\{\mathcal{M}_i^{k-1}\right\}$. If the retrieved target instance overlaps with the new instance by an Intersection-over-Union (IOU) greater than 0.5, the original record is replaced by $(f_{i,j}^k, c_{i,j}^k)$, otherwise, the new instance is directly inserted into the memory bank. This updating mechanism effectively eliminates outdated or unreliable records while preserving the latest pseudo-label information with high confidence.

\subsubsection{Geometry-Aware Confidence Calibration}

Wake often manifest as elongated local linear structures, and semantically, similar wake instances tend to be close to each other in the feature space. However, directly relying on feature similarity for pseudo-label calibration is easily disturbed by sea clutter, ship residual shadows, and other interferences. To address this, we design a Geometry-Aware Confidence Calibration mechanism, which explicitly characterizes the local geometric consistency of candidate regions while leveraging semantic similarity, thereby enhancing the discriminative capability for wakes. Specifically, let the feature-confidence memory bank at the $k$-th training epoch store $n_m$ target instances, $\left\{\mathcal{M}_i^{k}\right\} =\left\{\left(f_i, c_i\right)\right\}_{i=1}^{n_m}$, where $f_i$ and $c_i$ denote the feature representation and confidence of the $i$-th instance, respectively. For a new target domain sample $Z$, we obtain its feature representation $f_z$ and predicted confidence $c_z$ using the momentum model. For the $z$-th target instance, we compute the cosine similarity between its feature and all stored target features in the memory bank $\left\{\mathcal{M}_i^k\right\}$, and select the top-$K$ nearest neighbors, obtaining the corresponding neighbor index set $\mathcal{N}i$ and similarity set $\left\{\alpha_1, \ldots, \alpha_K\right\}$. Based on this, we define the weighted neighbor voting weight as
\begin{equation}
  w_i=\frac{\exp \left(\gamma \cdot \alpha_i\right)}{\sum_{j=1}^K \exp \left(\gamma \cdot \alpha_j\right)}
  \end{equation}
where $\gamma$ is a temperature coefficient. Next, we retrieve the corresponding neighbor confidences $\left\{c_{i_1}, c_{i_2}, \ldots, c_{i_k}\right\}$ from the memory bank and compute their weighted average
\begin{equation}
  c_{\text {neighbor}}=\sum_{j=1}^K w_j c_{i_j}
\end{equation}

Subsequently, this neighborhood confidence is fused with the target instance's own confidence $c_z$, yielding the neighborhood-fused confidence
\begin{equation}
  c_{f}=\delta \cdot c_z+(1-\delta) \cdot c_{\text {neighbor }}
\end{equation}
where $\delta$ is the fusion coefficient.

On this basis, we further compute the geometric consistency factor of the wake candidate region. For the feature $f_z$ of the target instance, the local structure tensor $J$ is obtained via gradient computation
\begin{equation}
J=\left[\begin{array}{cc}
\sum\left(\partial_x f_z\right)^2 & \sum\left(\partial_x f_z\right)\left(\partial_y f_z\right) \\
\sum\left(\partial_x f_z\right)\left(\partial_y f_z\right) & \sum\left(\partial_y f_z\right)^2
\end{array}\right] 
\end{equation}
Its eigenvalues $\lambda_1 \geq \lambda_2$ characterize the principal orientation and elongation of the region. To assess whether the candidate region exhibits a distinct linear structure, we introduce the anisotropy measure,
\begin{equation}
C=\frac{\lambda_1-\lambda_2}{\lambda_1+\lambda_2}
\end{equation}
However, anisotropy alone is insufficient to suppress non-structural noise. Therefore, we further adopt the vesselness measure
\begin{equation}
V=\left(1-\exp \left(-\frac{R_B^2}{2 \beta^2}\right)\right) \cdot \exp \left(-\frac{S^2}{2 c^2}\right)
\end{equation}
where $R_B={\left|\lambda_2\right|} / {\left|\lambda_1\right|}$ and $S=\sqrt{\lambda_1^2+\lambda_2^2}$. We then combine the two to obtain the geometry-aware factor.
\begin{equation}
c_{g}=\gamma_1 \bar{C}+\gamma_2 \bar{V}
\end{equation}
where $\bar{C}, \bar{V}$ denote the local averages within the pseudo-label target region, and $\gamma_1,\gamma_2$ are weighting hyperparameters. Intuitively, a larger geometry-aware factor indicates that the candidate region is more consistent with the linear geometric prior of wakes.

Finally, we fuse the neighborhood-fused confidence with the geometry-aware factor into the refined confidence of the instance
\begin{equation}
c^{\prime}=\mu c_{f}+(1-\mu) \sigma\left(c_{g}\right)
\end{equation}
where $\mu$ controls the fusion weight between neighborhood and geometry, and $\sigma(\cdot)$ is a normalization function.

In the process of pseudo-label calibration, the threshold setting directly determines the quality of pseudo-labels and the stability of training. To this end, we design a Global-Instance Adaptive Thresholding strategy. This strategy first employs Exponential Moving Average (EMA) \cite{wangfreematch} to gradually smooth the threshold updates, adapting to the model's evolving capability of representing target-domain features at different training stages. Specifically, the global threshold at the $k$-th epoch is defined as
\begin{equation} 
  \tau_k=\lambda \cdot \tau_{k-1}+(1-\lambda) \cdot \frac{1}{N} \sum_{b=1}^{N} \max \left(\psi_b\right),
\end{equation}
where $\alpha$ is the EMA decay factor, $N$ denotes the number of batches, and $\psi_b$ represents the maximum confidence score of detection boxes in the $b$-th batch. This design ensures that the threshold gradually increases as training progresses, transitioning from a lower bar in the early stage to a higher requirement in the later stage.

On this basis, we further introduce an instance-level neighborhood consistency adjustment mechanism to assign personalized retention thresholds for different candidate objects. Specifically, let the pseudo-label instance be $b_i$ with confidence $c_i$. From the FCMB, we retrieve $K$ similar neighbors ${c_j}_{j=1}^K$, and define the confidence dispersion as
\begin{equation}
\Delta_i=\frac{1}{K} \sum_{j=1}^K\left|c_i-c_j\right|
\end{equation}

Finally, the global-instance adaptive threshold is given by
\begin{equation}
\hat{\tau_k}=\tau_{k}+\eta \cdot \Delta_i
\end{equation}
where $\eta$ is the adjustment coefficient. If $c^{\prime}$ falls below the dynamic threshold $\hat{\tau_k}$, the candidate box is determined as background or low-confidence wake and discarded; otherwise, the updated confidence $\hat{c}=c^{\prime}$ is written back into the detection results.

\begin{table*}[t]
  \centering
  \caption{Experimental results from optical wakes to sar wakes. We use CSP-Darknet53 as the backbone network. The mean average precision (\%) is evaluated on the target images.}
  \label{tab:comparison}
      \begin{tabular*}{\textwidth}{@{\extracolsep{\fill}} c c c c c }
      \toprule
      \textbf{Method} & \textbf{Detector} & \textbf{Backbone} & \textbf{mAP@0.5} & \textbf{mAP@0.5:0.05:0.95} \\
      \midrule
      Source only      & YOLOv5       & CSP-Darknet53 & 20.22 &  7.57 \\
      \midrule
      DA Faster R-CNN\cite{chen2018domain}  & Faster R-CNN & VGG-16        & 32.00 &  9.00 \\
      PT\cite{chen2022learning}               & Faster R-CNN & VGG-16        & 29.02 &  9.51 \\
      CMT-PT\cite{cao2023contrastive}           & Faster R-CNN & VGG-16        & 23.12 &  6.47 \\
      PMT\cite{belal2024multi}          & Faster R-CNN       & VGG-16  & 31.87 & 11.58 \\
      EMPs\cite{hsu2020every}             & FCOS         & VGG-16        & 25.90 &  7.70 \\
      SCAN\cite{li2022scan}             & FCOS         & VGG-16        & 31.99 & 10.20 \\
      SCAN++\cite{li2022scan++}           & FCOS         & VGG-16        & 35.99 & 11.32 \\
      SIGMA\cite{li2022sigma}            & FCOS         & VGG-16        & 44.03 & 15.50 \\
      MTM\cite{weng2024mean}            & Deformable DETR         & ResNet-50        & 35.94 & 13.26 \\
      ConfMix\cite{mattolin2023confmix}          & YOLOv5       & CSP-Darknet53 & 45.11 & 13.81 \\
      VT\cite{yang2025versatile}               & YOLOv5       & CSP-Darknet53 & 35.45 & 12.64 \\
      SimMemDA (Ours)         & YOLOv5       & CSP-Darknet53 & \textbf{57.03} & \textbf{19.65} \\
      \midrule
      Oracle           & YOLOv5       & CSP-Darknet53 & 66.97 & 27.20 \\
      \bottomrule
      \end{tabular*}
\end{table*}

\subsection{Reliable Region Selection and Sample Mixing}
Inspired by \cite{mattolin2023confmix}, to fully leverage the annotated information in source domain data and the calibrated pseudo-labels from target domain data, we adopt a region-level mixing strategy that integrates source and target domain data, thereby enabling mixed-domain supervised learning. Specifically, we first randomly sample one source domain image $x_i^S$ and one target domain image $x_j^T$. Next, we calibrate the pseudo-labels for the target domain image to obtain its prediction results, and subsequently partition this image into four equal regions. Then, we calculate the average pseudo-label confidence within each region, treating this value as the region's confidence indicator. Finally, we select the region with the highest confidence and blend it with the source domain data to generate mixed data
\begin{equation}
  x_M=M^T \odot x_i^S+\left(1-M^T\right) \odot x_j^T,
  \end{equation}
where $M^T$ denotes the mask corresponding to the selected region.

\subsection{Adaptive detector training}
To facilitate the training of the adaptive detector, we employ two main loss functions: supervised detection loss $\mathcal{L}_{det}$ and self-supervised consistency loss $\mathcal{L}_{cons}$. In particular, $\mathcal{L}_{det}$ preserves the domain-specific knowledge acquired from the source domain task during adaptation, while $\mathcal{L}_{cons}$ encourages the target domain features to gradually adapt to the target task by penalizing discrepancies between predictions on mixed images and the combined pseudo-labels. 
To construct the mixed images, we combine the predicted instances from the source domain with calibrated pseudo-labels from the target domain. Let $\left\{b_i^T\right\}$ denote the calibrated pseudo-label instances in the target domain, and $\left\{b_i^S\right\}$ the current predictions from the detector $G$ on source domain data. We select a subset of target pseudo-labels located within a predefined target region, denoted as $\left\{b_i^T\right\}^R$, and merge them with source domain predictions outside this region, denoted as $\left\{b_i^S\right\}^{-R}$, to form the set of combined pseudo-labels $\left\{b_i^{S, T}\right\}$. For bounding boxes that exceed the boundaries of the defined region, we crop them to ensure they remain within bounds. After forwarding the mixed images through the detector, we obtain the set of predicted instances, denoted as $\left\{b_i^M\right\}$. The supervised detection loss and the self-supervised consistency loss are then defined as:
\begin{equation}
  \mathcal{L}_{det}=\mathcal{L}\left(\left\{b_i^S\right\}, \{y_i^S\}\right) ,
   \mathcal{L}_{cons}=\mathcal{L}\left(\left\{b_i^M\right\},\left\{b_{i}^{S, T}\right\}\right),
  \end{equation}
where ${y_i^S}$ represents the ground-truth annotations for the source domain, and $\mathcal{L}(\cdot)$ denotes the specific form loss function. In the case of one-stage detectors such as YOLOv5, this loss function typically includes three components: the bounding box regression loss $L_{box}$, the object confidence loss $L_{obj}$, and the classification loss $L_{cls}$.

Ultimately, the total loss function is defined as the weighted sum of the two losses
\begin{equation}
  \mathcal{L}_{\mathrm{total}}=\mathcal{L}_{d e t}+\alpha \mathcal{L}_{cons},
\end{equation}
where the hyperparameter $\alpha$ is used to balance the weights between the supervised loss and the self-supervised consistency loss. When pseudo-labels from the target domain are more reliable, a higher weight should be assigned to $\mathcal{L}_{cons}$; otherwise, the weight should be reduced. To this end, we design a dynamic weighting method that adjusts $\alpha$ by smoothly evaluating the reliability of each pseudo-label in the mixed images. Specifically, for each pseudo-label $b_{i}^{S, T}$ in the mixed image, with confidence $c_i$, we adopt a soft-thresholding mechanism such that its contribution to the weight is continuous rather than based on a simple hard-threshold decision. The calculation formula for the dynamic weight $\alpha$ is
\begin{equation}
  \alpha=\frac{1}{n} \sum_{i=1}^n \sigma\left(\kappa\left(c_{i}-c_{\mathrm{th}}\right)\right),
  \end{equation}
where $\sigma(z)=\frac{1}{1+e^{-z}}$ is the standard sigmoid function, and $\kappa$ is a scaling factor controlling the difference between confidence and the preset threshold $c_{\mathrm{th}}$.

\begin{figure*}
  \centering 
  \includegraphics[width=\textwidth]{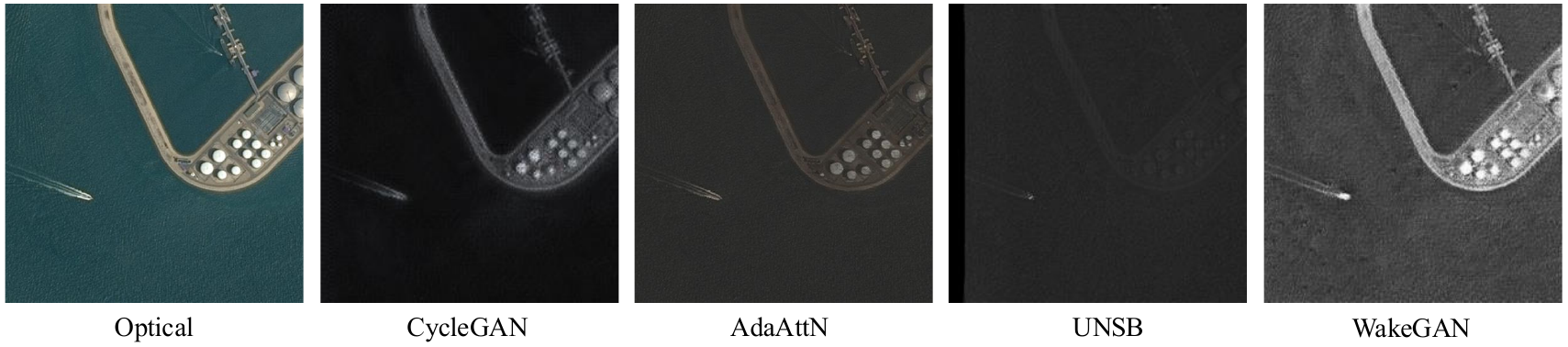} 
  \caption{Visual comparison of optical to SAR style transfer results. From left to right: input optical image, and the translated results obtained by CycleGAN, AdaAttN, UNSB, and the proposed WakeGAN.} 
  \label{fig:style_transfer} 
\end{figure*}

\section{Experiments}\label{section_4}
\subsection{Datasets and Evaluation Metrics}
This section evaluates the adaptation performance of the proposed SimMemDA method on the SAR ship wake dataset. The dataset used comprises the optical ship wake dataset SWIM \cite{xue2021rethinking} and the SAR ship wake dataset OpenSARWake \cite{xu2024opensarwake}, specifically described as follows:
\begin{itemize}
   \item \emph{SWIM} \cite{xue2021rethinking}: This dataset is the first publicly available optical ship wake detection dataset known so far. It consists of 14,610 optical satellite and aerial images, with annotations for 15,356 wake instances, all of which have been re-labeled with horizontal bounding boxes. The dataset covers Kelvin wakes and turbulent wakes generated by various types of vessels ranging from yachts to large container ships. Images were primarily obtained from Google Earth, spanning coastal areas in Asia, Europe, Africa, North America, and Oceania, collected from 2009 to 2021. The images have spatial resolutions ranging from 0.5 to 2.5 meters and have been uniformly cropped to $512 \times 512$ pixels. They encompass diverse sea states, lighting, weather, and imaging conditions, ensuring rich intra-class variations among wake instances.
   \item \emph{OpenSARWake} \cite{xu2024opensarwake}: This dataset is the first publicly available large-scale SAR ship wake dataset, comprising 3,973 SAR images across L-, C-, and X-bands, with annotations for 4,096 wake instances, which have also been re-labeled using horizontal bounding boxes. Images in this dataset primarily originate from three mainstream satellite SAR sensors, namely Sentinel-1A, TerraSAR-X, and ALOS-PALSAR, covering both Kelvin wakes and turbulent wakes under various sea states and imaging modes.
 \end{itemize}

In the target domain, we employ mean average precision (mAP) as the primary evaluation metric. Average precision (AP) is derived by comprehensively computing precision and recall for each category at various confidence thresholds. Specifically, precision (P) and recall (R) are formally defined as:
\begin{equation}
  P=\frac{\mathrm{TP}}{\mathrm{TP}+\mathrm{FP}}, \quad R=\frac{\mathrm{TP}}{\mathrm{TP}+\mathrm{FN}},
  \end{equation}
where $\mathrm{TP}$,$\mathrm{FP}$ and $\mathrm{FN}$ represent true positives, false positives, and false negatives, respectively. Subsequently, the AP of each category is obtained by plotting and integrating the Precision-Recall (P-R) curve across various thresholds, formally expressed as:
\begin{equation}
  \mathrm{AP}=\int_0^1 P(R) d R.
  \end{equation}

During evaluation, an Intersection over Union (IoU) threshold commonly utilized in object detection is applied, whereby a detection is considered a true positive only if the IoU between the predicted box and ground-truth box exceeds this predefined threshold. The mAP is then computed by averaging the AP values across all categories, formally expressed as:
\begin{equation}
  \mathrm{mAP}=\frac{1}{N_c} \sum_{c=1}^{N_c} \mathrm{AP}_c,
  \end{equation}
where ${N_c}$ denotes the total number of categories. The mAP metric provides a comprehensive evaluation of detection performance across multiple categories, balancing the influence of precision and recall, thus serving as an effective and widely adopted metric in multi-class object detection tasks

\begin{figure*}[!t]
  \centering
  \subfloat[]{\includegraphics[width=0.24\textwidth]{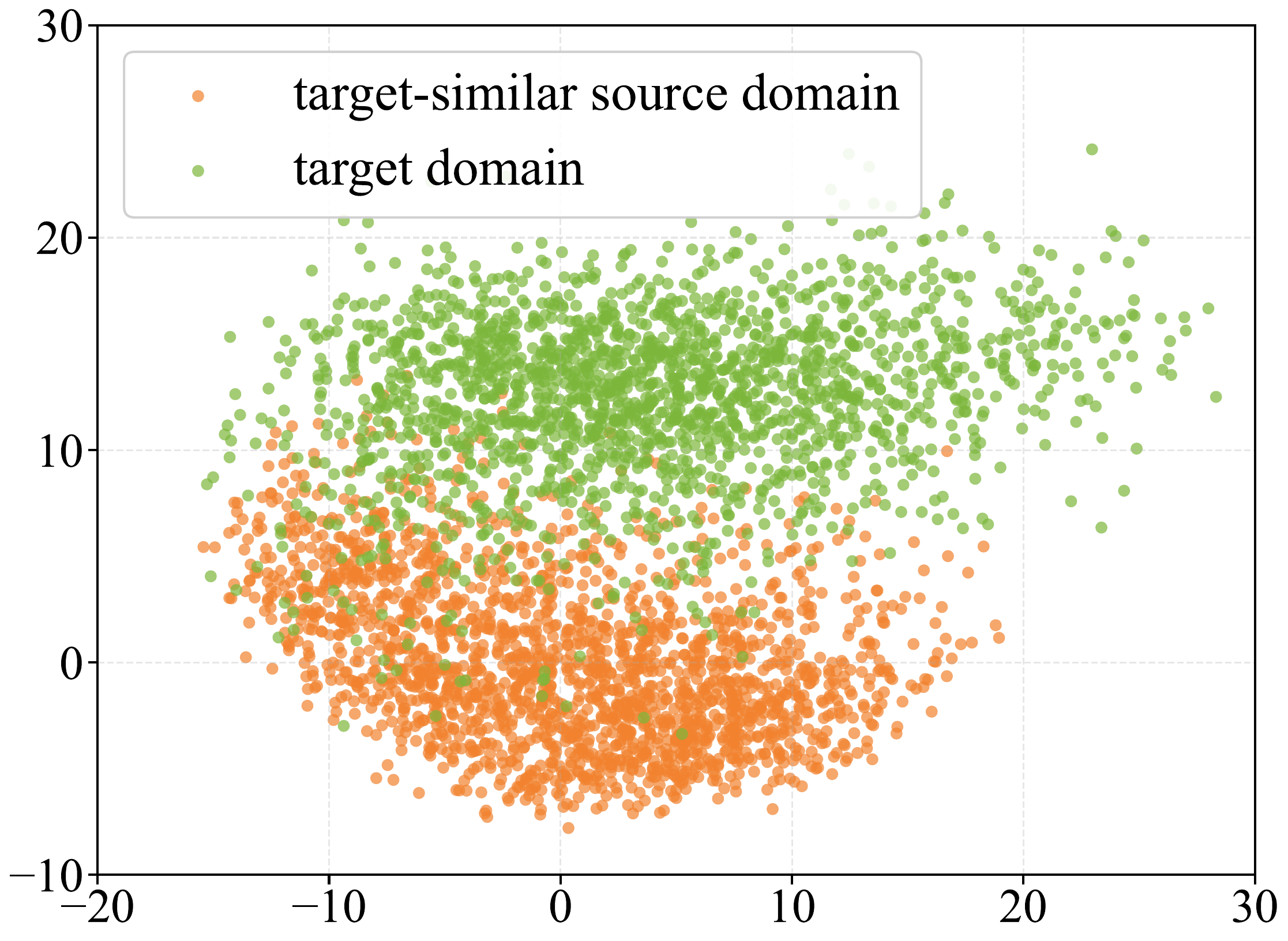}\label{fig:pca_similar_vs_sar}}
  \hfill
  \subfloat[]{\includegraphics[width=0.24\textwidth]{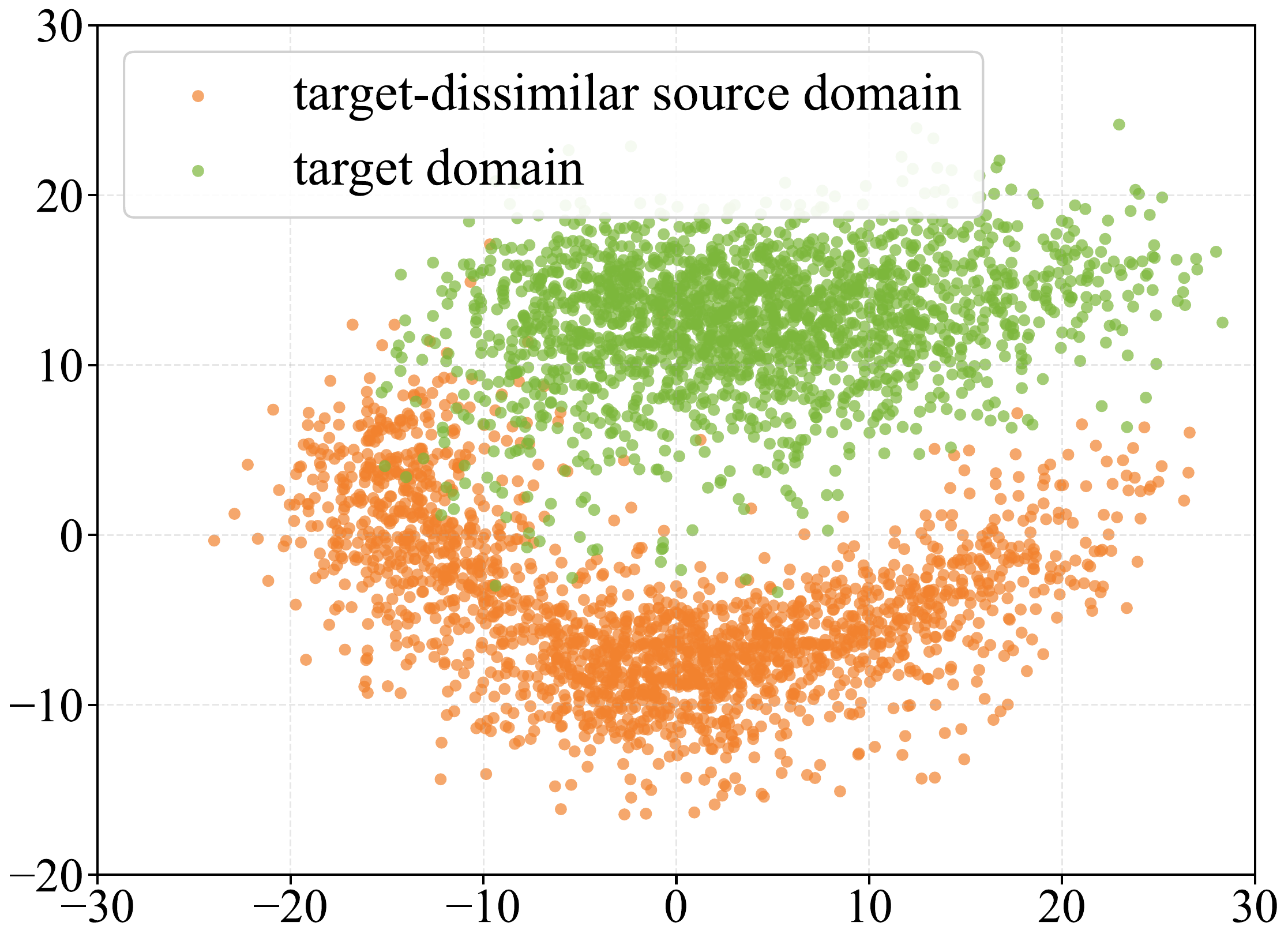}\label{fig:pca_dissimlar_vs_sar}}
  \hfill
  \subfloat[]{\includegraphics[width=0.24\textwidth]{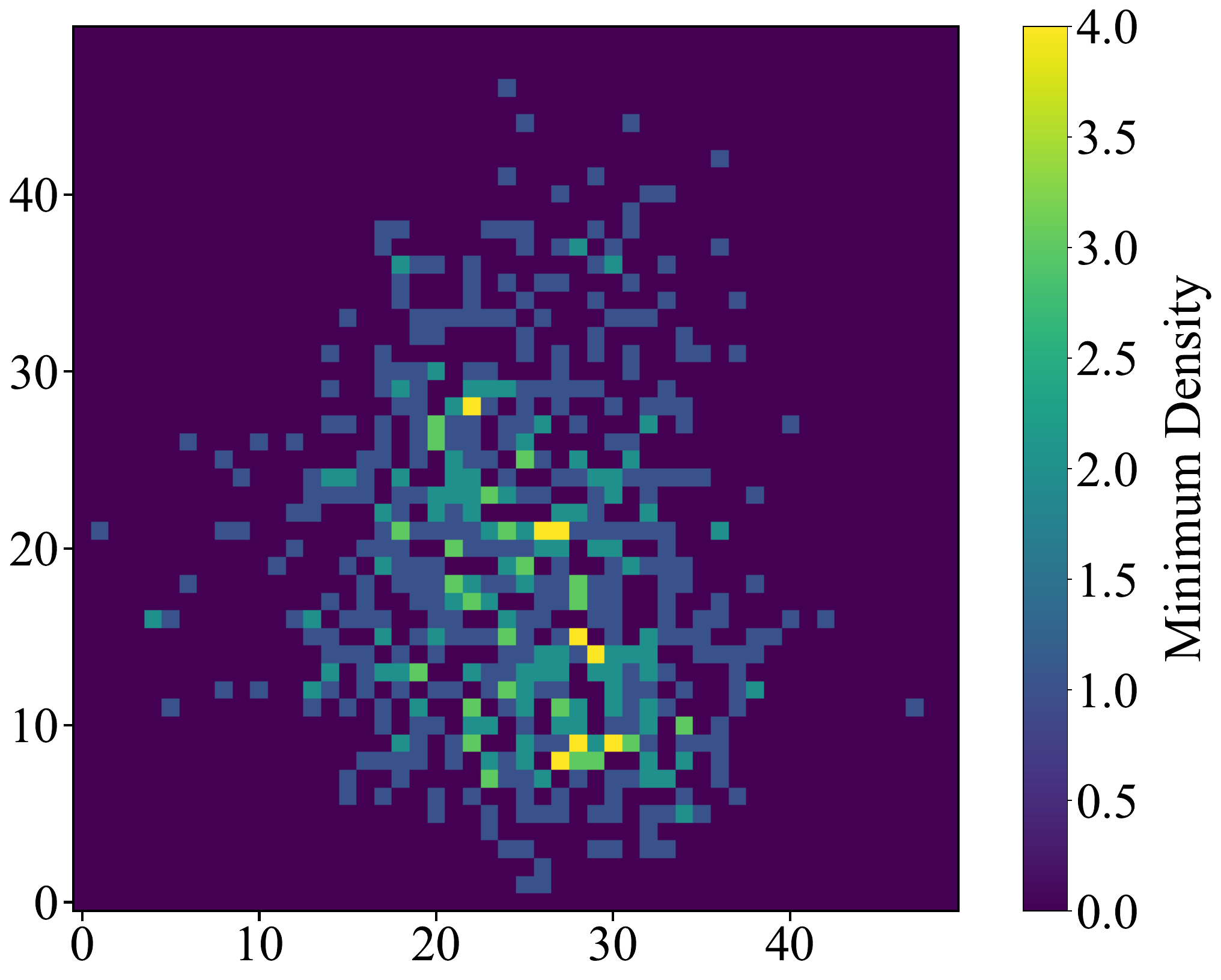}\label{fig:density_simlar_vs_sar}}
  \hfill
  \subfloat[]{\includegraphics[width=0.24\textwidth]{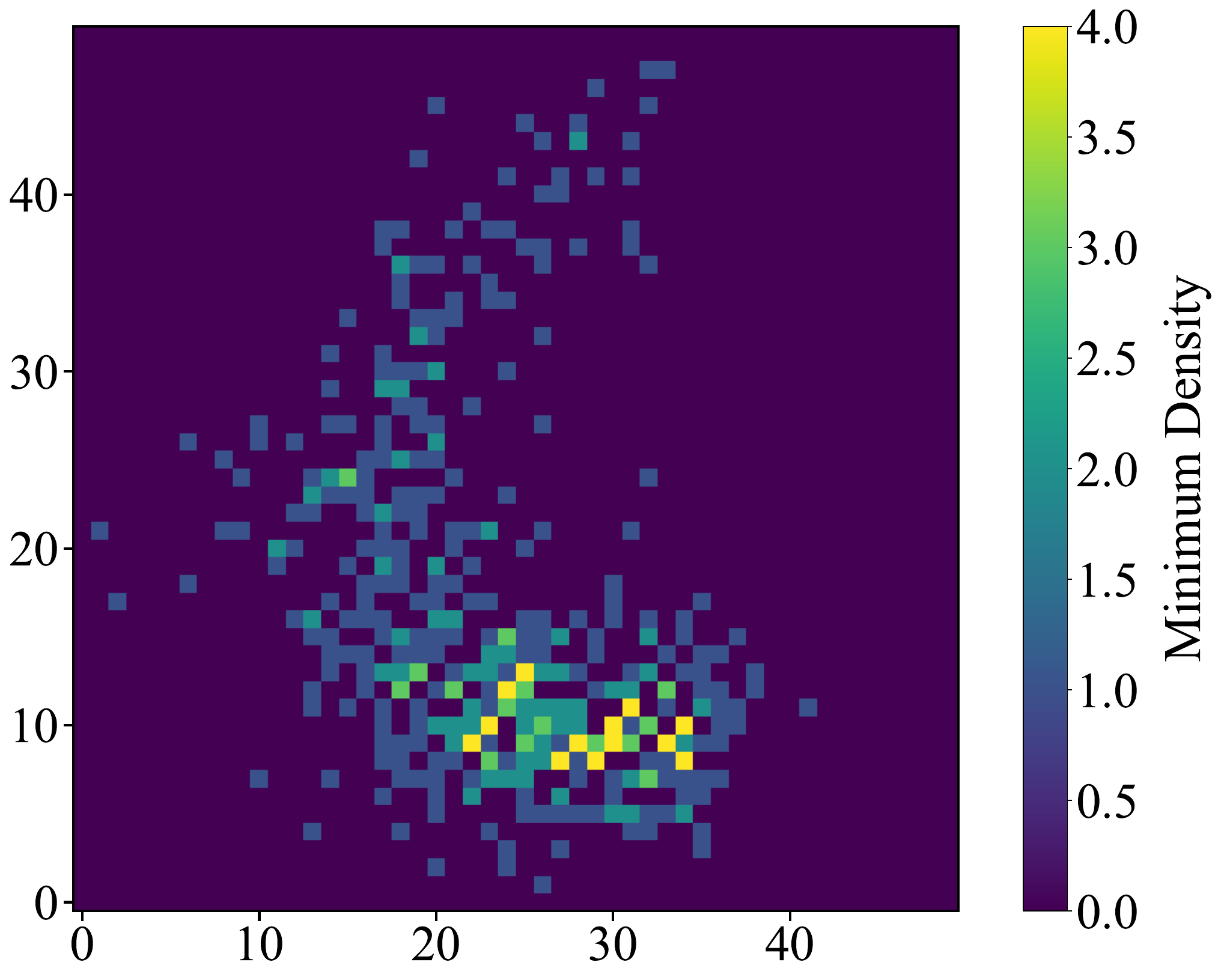}\label{fig:density_dissimlar_vs_sar}}
  
  \caption{The scatter plots in (a) and (b) depict the distribution of data in the PCA-reduced space, while the density overlap heat maps in (c) and (d) quantify inter-domain distributional similarity, with higher intensity colors representing greater overlap. (a) presents the distribution of target-similar source domain data alongside target domain data in the two-dimensional dimensionality-reduced space. Correspondingly, (b) illustrates the distribution of target-dissimilar source domain data relative to the target domain data. (c) displays the density overlap heat map for target-similar source domain, exhibiting a density overlap rate of 31.44\%, whereas (d) presents the density overlap heat map for target-
dissimilar source domain, demonstrating a lower density overlap rate of 23.22\%. }
  \label{fig:similar_dissimilar}
\end{figure*}

\begin{table}[t]
  \centering  
  \caption{The core component ablation studies of the proposed method, which uses CSP-Darknet53 as the backbone network. The mean average precision (\%) is evaluated on the target images. Domain Style Transfer Sec. \ref{section_31}, Similarity-Guided Filtering Sec. \ref{section_32}, Pseudo-Label Calibration Sec. \ref{section_33}.}
  \label{tab:ablation}
  \resizebox{\linewidth}{!}{
  \begin{tabular}{cccccc}  %
    \toprule
    \textbf{\makecell{Method}} & \textbf{\makecell{Domain\\Style Transfer}} & \textbf{\makecell{Similarity-\\Guided Filtering}} & \textbf{\makecell{Pseudo-Label\\Calibration}} & \textbf{\makecell{mAP@0.5}} & 
    \textbf{\makecell{mAP@0.5:\\0.05:0.95}} \\
    \midrule
    Source Only & & & & 20.22 & 7.57 \\
    \midrule
    \multirow{3}{*}{SimMemDA} & $\checkmark$& & & 41.35 & 13.87 \\
     & $\checkmark$& $\checkmark$& & 46.61 & 14.83 \\
     & $\checkmark$& $\checkmark$& $\checkmark$& 57.03 & 19.65 \\
    \bottomrule
  \end{tabular}
  } %
\end{table}

\subsection{Implementation Details}
In this experiment, we selected the lightweight YOLOv5s architecture from the YOLOv5 series as our detection model, employing CSPDarknet53 as the backbone for feature extraction. The batch size was set to 16, comprising 8 source domain images and 8 target domain images per batch, with each image uniformly resized to $512 \times 512$ pixels. Initially, the model underwent pre-training on an optical ship wake dataset, initialized with weights from the COCO pre-trained model. Subsequently, adaptive learning was performed. In the Non-Maximum Suppression stage, we set the IoU threshold to 0.5 and the confidence threshold to 0.25 to generate pseudo-detection results. When applying exponential moving average for smoothing threshold updates, the parameter $\lambda$ was set to 0.9, and the initial threshold $\tau_{0}$ was 0.05. In the computation of $\alpha$, the confidence threshold $c_{th}$ was set to 0.5, with a scaling factor $\kappa$ of 10. All experiments were conducted using the PyTorch framework.

\subsection{Result}

In this section, We compared the proposed SimMemDA method against current state-of-the-art unsupervised domain adaptation methods using a ship wake dataset. Specifically, we select the following categories of methods for comparison: domain alignment learning methods such as DA Faster R-CNN\cite{chen2018domain}, EMPs\cite{hsu2020every}, SCAN\cite{li2022scan}, SCAN++\cite{li2022scan++} and SIGMA\cite{li2022sigma}; pseudo-label methods such as ConfMix\cite{mattolin2023confmix} and VT\cite{yang2025versatile}; and auxiliary model strategies such as PT\cite{chen2022learning} , CMT-PT\cite{cao2023contrastive}, PMT\cite{belal2024multi} and MTM\cite{weng2024mean}. Additionally, we introduced a Source Only model trained solely on labeled optical wake data and evaluated on SAR wake data as a lower bound, and an Oracle model trained and evaluated exclusively on SAR data as an upper bound. To facilitate comparison with existing methods, we adopt pseudo-SAR data as the source domain and SAR data as the target domain. Traditionally, comparative methods utilize optical data as the source domain and SAR data as the target domain. However, in our method, we introduce an additional processing step to transform optical data into pseudo-SAR data to reduce domain differences between optical and SAR data. To avoid any impact of this data processing approach on the experimental results, we similarly transform the original optical source domain data into pseudo-SAR data for the comparative experiments, ensuring that all methods are evaluated under unified data conditions. Specifically, all baseline methods were retrained from scratch with their official implementations and recommended optimal hyper-parameter settings, under the unified setting where pseudo-SAR images are used as the source domain and real SAR images as the target domain, ensuring fair and consistent comparisons. The results presented in Table \ref{tab:comparison} illustrate that the Source Only model achieved a low mAP@0.5 of 20.22\%, highlighting substantial domain discrepancies. In contrast, our SimMemDA method achieved an mAP@0.5 of 57.03\%, surpassing the previously best-performing method by 11.92\%. Under the stricter metric mAP@0.5:0.05:0.95, SimMemDA also attained the highest score of 19.65\%, outperforming all comparison methods. These outcomes demonstrate the effectiveness of SimMemDA in reducing domain discrepancies and enhancing both localization and classification accuracy in cross-domain detection tasks.

\begin{table}[htbp]
  \centering
  \caption{Comparison of different style transfer methods for optical to SAR wake detection.}
  \begin{tabular}{ccc}
    \toprule
     \textbf{Style Transfer Methods} & \textbf{mAP@0.5} & \textbf{mAP@0.5:0.05:0.95}    \\
    \midrule
    CycleGAN \cite{zhu2017unpaired}  & 52.45 & 17.67 \\
    AdaAttN \cite{liu2021adaattn}   & 50.27 & 16.86 \\
    UNSB  \cite{kim2024unpaired}     & 39.79 & 12.17 \\
    WakeGAN    & \textbf{57.03} & \textbf{19.65} \\
    \bottomrule
  \end{tabular}
  \label{tab:style_transfer}
\end{table}

\begin{figure}[!t]
  \centering
  \subfloat[]{%
    \includegraphics[width=0.5\columnwidth]{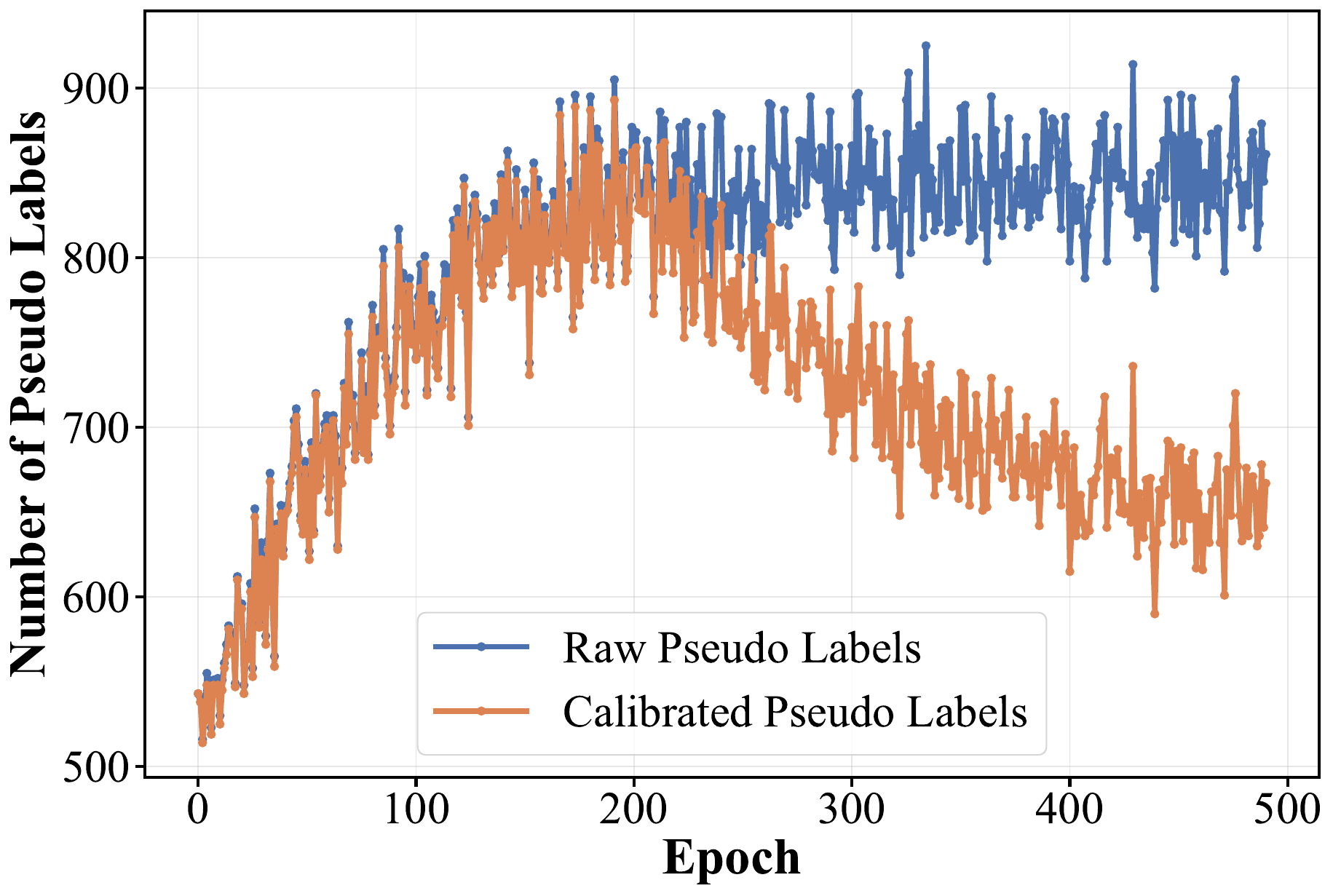}%
    \label{fig:box_count_trend}
  }
  %
  \subfloat[]{%
    \includegraphics[width=0.5\columnwidth]{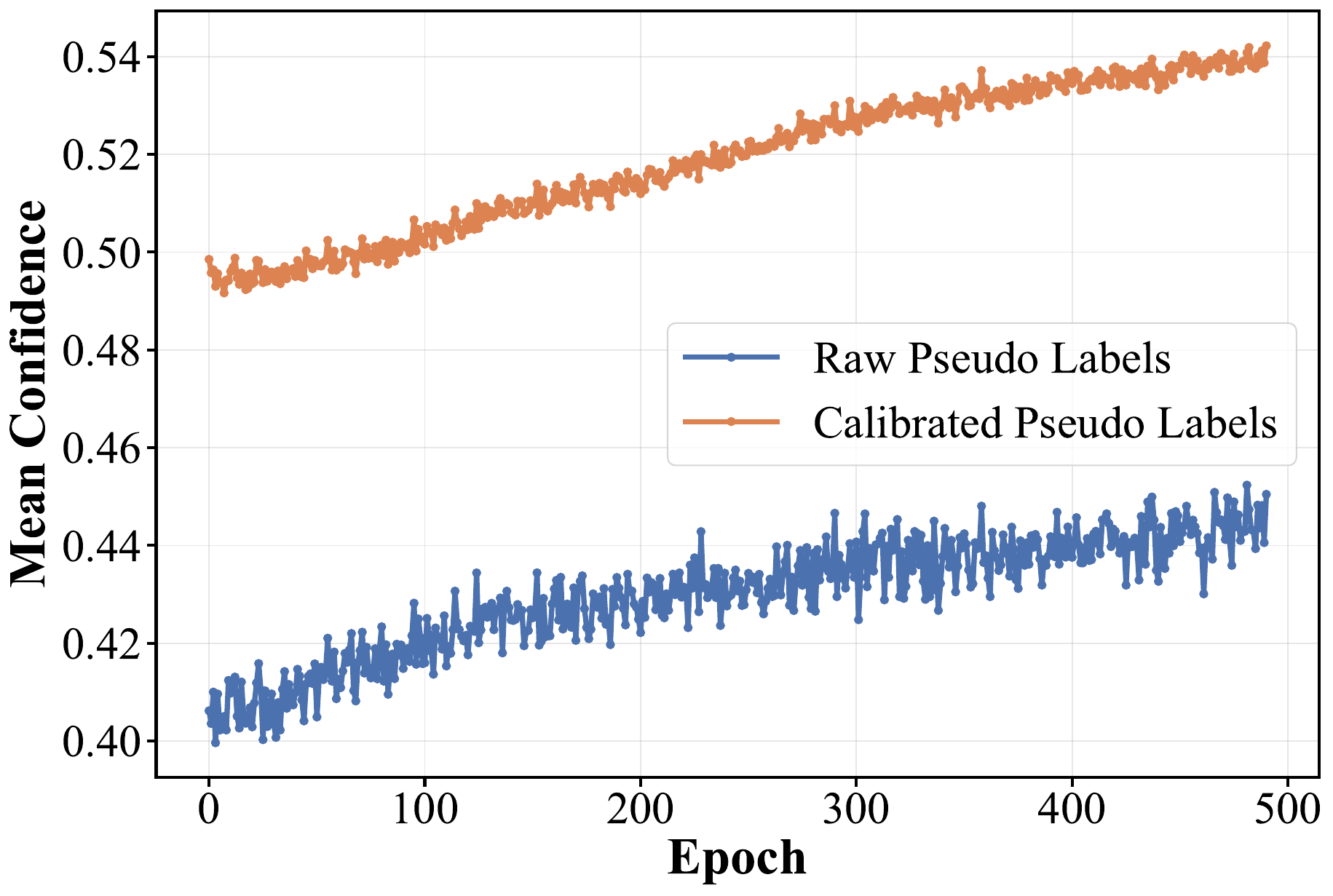}%
    \label{fig:mean_confidence_trend}
  }
  \caption{(a) shows the variation trend of the number of raw pseudo-labels (blue line) and calibrated pseudo-labels (orange line) by the memory-guided geometric-aware calibration during training; (b) presents the change in the mean confidence of raw pseudo-labels (blue line) and calibrated pseudo-labels (orange line) memory-guided geometric-aware calibration during training.}
  \label{fig:box_count_and_confidence}
\end{figure}

\begin{figure}[htbp]
  \centering 
  \includegraphics[width=\columnwidth]{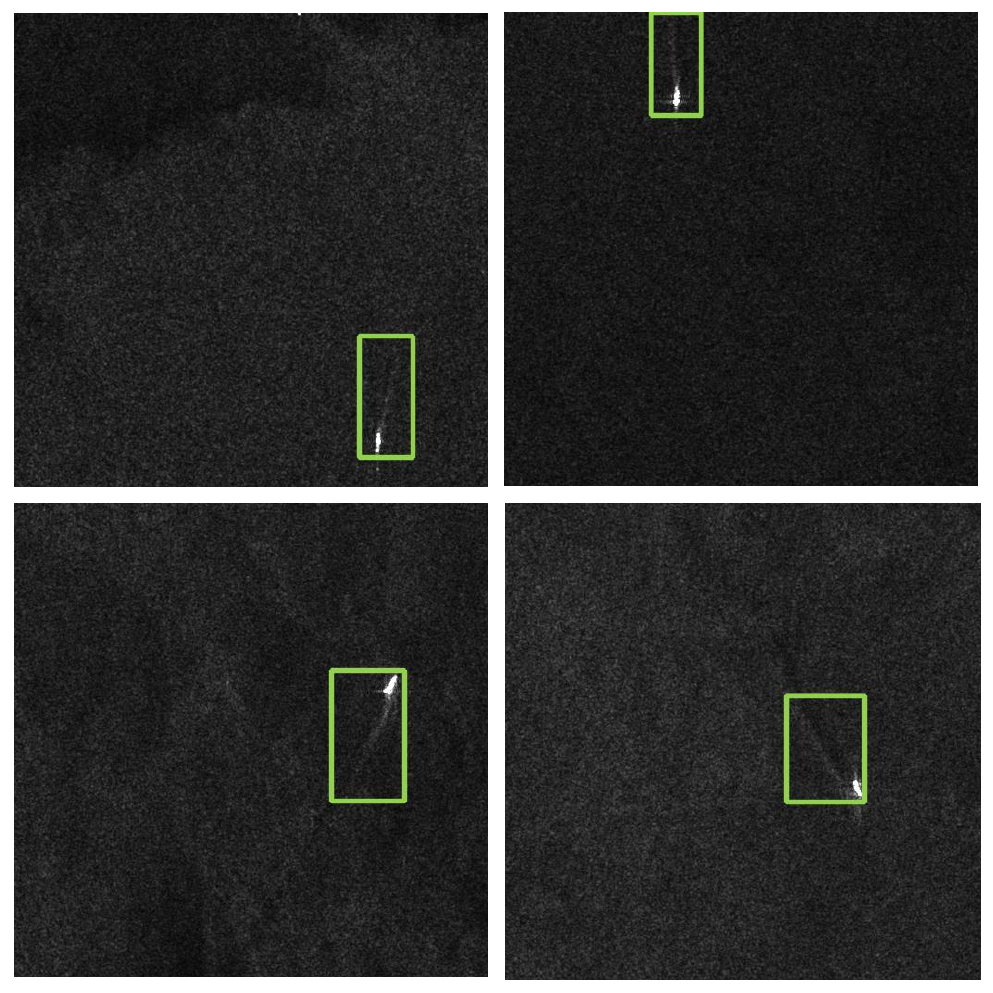} 
  \caption{SimMemDA ship wake detection results in the SAR imagery from the Sentinel-1 satellite.} 
  \label{fig:sentinel_detect} 
\end{figure}

\subsection{Ablation Studies}

To evaluate the contributions of the core components proposed in our method, we conduct ablation experiments. The specific results are shown in Table~\ref{tab:ablation}. We observe that introducing structure-preserving domain style transfer improves the mAP@0.5 and mAP@0.5:0.95 of the Source Only baseline to 41.35\% and 13.87\%, respectively, which indicates that generating pseudo-SAR images through WakeGAN alleviates the visual discrepancies between optical and SAR images. Building on this, further adopting similarity-guided source data filtering, which retains only the samples similar to the target domain distribution, raises mAP@0.5 to 46.61\% and mAP@0.5:0.95 to 14.84\%. This demonstrates that eliminating noisy samples helps mitigate negative transfer. Finally, by incorporating memory-guided geometry-aware pseudo-label calibration, the detection performance is further improved, achieving mAP@0.5 of 57.03\% and mAP@0.5:0.95 of 19.65\%. This shows that leveraging neighborhood feature consistency and wake-line geometry priors enhances the quality of pseudo labels on unlabeled target domain data, which benefits mixed-sample adaptive training and effectively strengthens the robustness of cross-domain detection.

\begin{figure*}
  \centering 
  \includegraphics[width=0.9\textwidth]{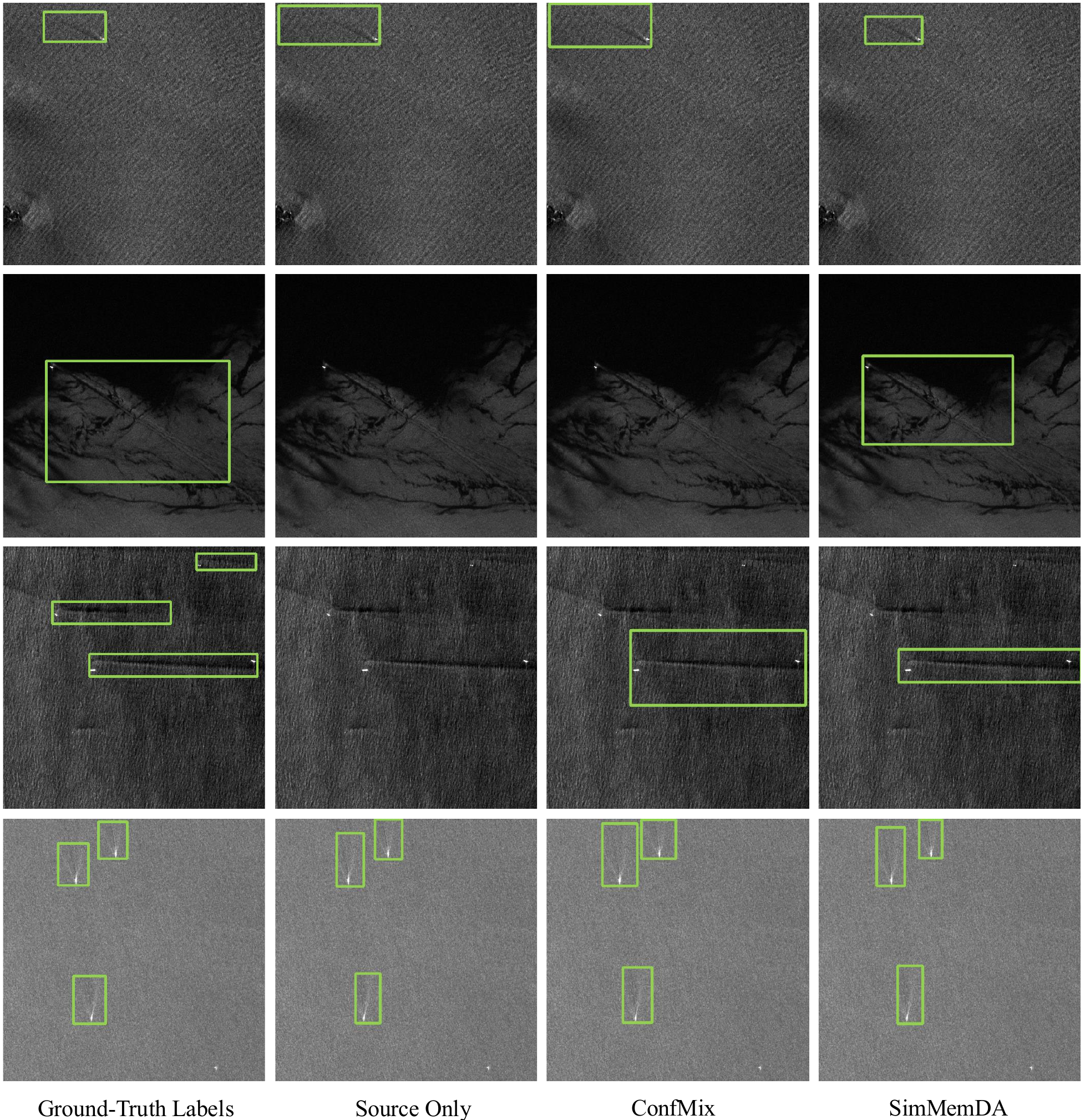} 

  \caption{Comparison of cross-domain SAR ship wake detection results. The left column shows the ground-truth labels, followed by the detection results from the Source Only, ConfMix, and SimMemDA methods, respectively. The SimMemDA method achieves more accurate bounding box localization. Not only does it more precisely match the ground-truth target boundaries in the first row, but it also successfully captures the complete ship wake in the second row, demonstrating a noticeably lower miss detection rate. } 
  \label{fig:detection} 
\end{figure*}

\begin{figure}[htbp]
  \centering 
  \includegraphics[width=\columnwidth]{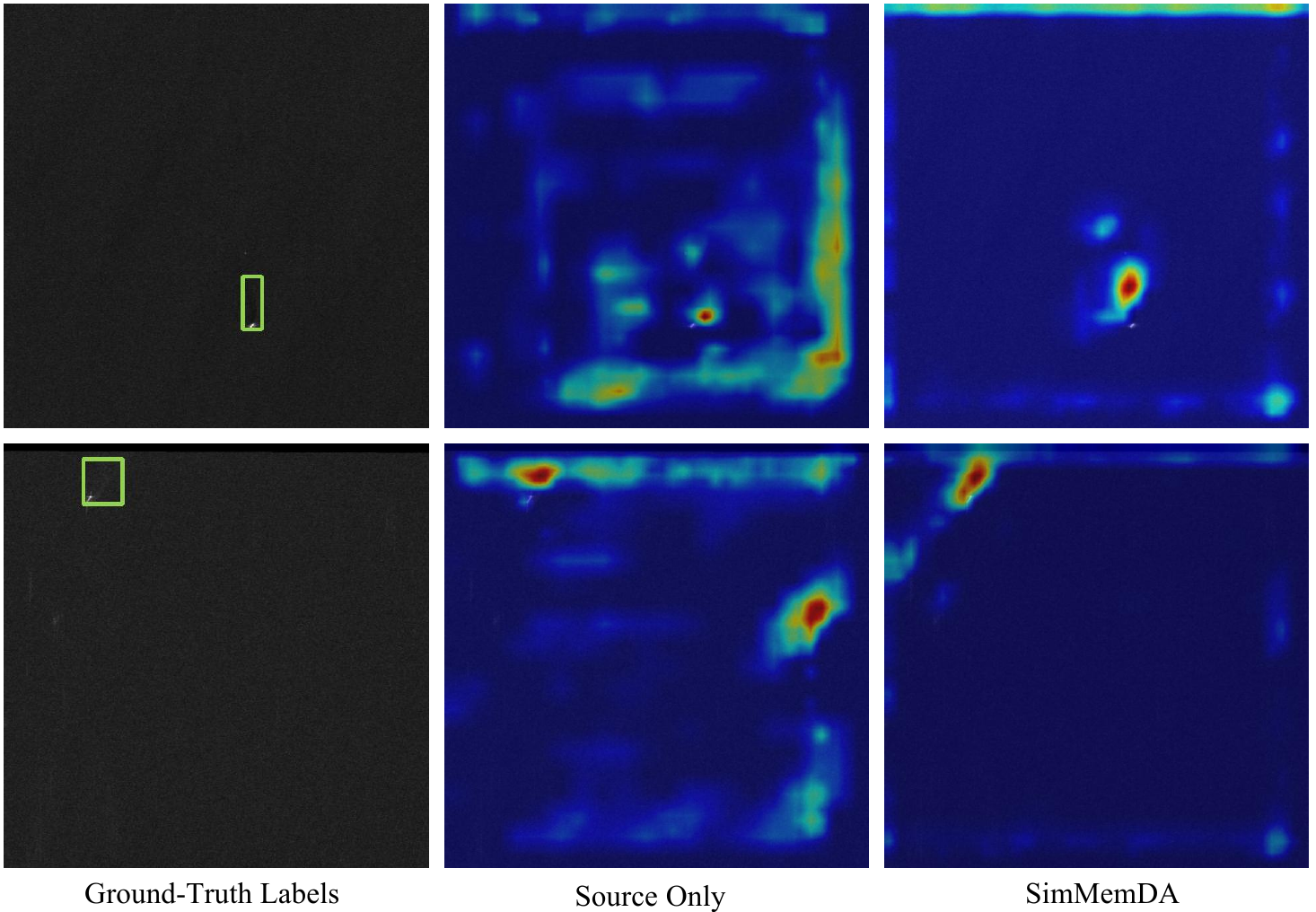} 
  \caption{SAR ship wake detection feature visualization comparison results. The left column shows the SAR images with ground-truth labels; the middle column shows the feature heatmaps generated by the Source Only; the right column shows the feature heatmaps generated by the SimMemDA.} 
  \label{fig:heatmap} 
\end{figure}

\subsection{Analysis of pseudo-SAR images}
In this section, we conduct a visual analysis of the pseudo-SAR images generated by WakeGAN and compare them with mainstream style transfer models, namely CycleGAN \cite{zhu2017unpaired}, AdaAttN \cite{liu2021adaattn}, and UNSB \cite{kim2024unpaired}. Fig. \ref{fig:style_transfer} illustrates the visual results of these four methods in the optical to SAR wake translation task. To further validate their effectiveness more intuitively, we use the generated pseudo-SAR samples as the source domain training set and evaluate detection performance on real SAR data from the target domain. The quantitative results are reported in Table \ref{tab:style_transfer}. From the comparison between visual effects and quantitative results, it can be observed that WakeGAN, under the dual constraints of structure preservation and detail enhancement, is able to simultaneously maintain consistency in both low-frequency background and high-frequency textures. This not only yields visually clear structures and realistic backgrounds but also significantly improves the performance of cross-domain detection models. This demonstrates that WakeGAN can effectively reduce the cross-domain distribution discrepancy and generate training samples that are more consistent with SAR domain characteristics. In contrast, although CycleGAN achieves a certain degree of overall style transfer, it shows limitations in wake detail preservation and noise suppression, resulting in only limited improvements in cross-domain detection performance. AdaAttN, by introducing adaptive attention, achieves some improvements in local texture transfer, but the wake contours remain blurry, performing worse than CycleGAN. UNSB, while relatively better in maintaining low-frequency background consistency, suffers from severe high-frequency structure loss, leading to blurred wake edges and more noise, providing the least benefit to cross-domain detection and ultimately yielding the worst detection performance. Overall, the results generated by WakeGAN surpass those of the compared methods in both visual realism and geometric structure preservation, especially in complex sea surface backgrounds where it can more stably reconstruct the physical characteristics of wakes.

\subsection{Feature Distribution Visualization of Target-similar and Target-dissimilar Source Images}
To further verify the effectiveness of target-similar source domain data filtering in aligning feature distributions, we visualized and compared features from target-similar source data, target-dissimilar source data, and target domain data using Principal Component Analysis (PCA)\cite{mackiewicz1993principal}.  After projecting the features onto a two-dimensional space, we quantitatively assessed distribution overlap. As illustrated in Fig. \ref{fig:similar_dissimilar}, target-similar source data showed significantly reduced distribution differences with a density overlap rate of 31.44\%, which is 8.11\% higher than the 23.22\% overlap rate of target-dissimilar source data. From the results of the feature distribution visualization, it is evident that the filtered source domain data exhibits a more pronounced overlap with the target domain data in the high-density regions, indicating that the selected target-similar source domain subset has greater potential for feature alignment, thereby helping to alleviate the distribution shift problem in the domain adaptation process.

\subsection{Memory-guided Pseudo-label Calibration Strategy Analysis}

In this section, to thoroughly analyze the effectiveness of the proposed memory-guided geometric-aware pseudo-label calibration strategy and the global-instance adaptive threshold strategy, we conduct the analysis from two perspectives: the evolution trend of pseudo-label quantity during training and the variation in the mean confidence of pseudo-labels. As illustrated in Fig. \ref{fig:box_count_trend} and Fig.\ref{fig:mean_confidence_trend}, Fig.\ref{fig:box_count_trend} shows that in the early training stage, the numbers of raw pseudo-labels and calibrated pseudo-labels remain nearly identical, enabling the model to fully exploit pseudo-labels for learning. However, after approximately 200 epochs, the number of calibrated pseudo-labels begins to decrease significantly. This reduction results from the introduction of the adaptive threshold strategy, which gradually removes low confidence and semantically ambiguous pseudo-labels, allowing the model in later training stages to rely primarily on high confidence pseudo-labels for optimization, thereby effectively reducing the interference of noisy supervision signals. Furthermore, Fig.\ref{fig:mean_confidence_trend} illustrates the trend of mean pseudo-label confidence throughout the training process. It can be observed that under the guidance of neighborhood consistency constraints from memory bank and geometric-aware calibration, the mean confidence of pseudo-labels after calibration is consistently higher and shows a faster growth trend beyond 200 epochs compared to raw pseudo-labels. This not only reflects the improvement in the model's detection capability but also indicates a continuous enhancement in the overall quality of pseudo-labels. In summary, the experimental results validate that memory-guided geometric-aware pseudo-label calibration achieves a favorable balance between pseudo-label quantity and quality, thereby providing more stable and high-quality supervision signals for cross-domain wake detection.

\subsection{Generalization Capability Validation}
To evaluate the generalization capability of the proposed method under complex maritime and varying imaging conditions, we selected SAR imagery acquired by Sentinel-1 in the Baltic Sea region and partitioned it into $512 \times 512$ pixel sub-images to construct the test set. The Baltic Sea exhibits significant spatiotemporal heterogeneity in wind fields, ocean currents, and sea surface roughness; as a result, backscatter intensities vary markedly under different seasonal and meteorological conditions, which poses challenges for detection algorithms due to interference from sea clutter, dark flows, and varying sea surface textures. As illustrated in Fig. \ref{fig:sentinel_detect}, our method successfully detects ship wakes in SAR data from unseen scenarios, with detection outcomes validated through matching with AIS data. Moreover, the SimMemDA framework consistently learns and represents wake features across diverse maritime conditions, accurately capturing the slender structures and arcuate shapes of ship wakes even amidst multiple prominent backscatter modes. This demonstrates the robust cross-domain adaptation capability of our approach in real-world maritime environments.

\begin{figure}[htbp]
  \centering 
  \includegraphics[width=\columnwidth]{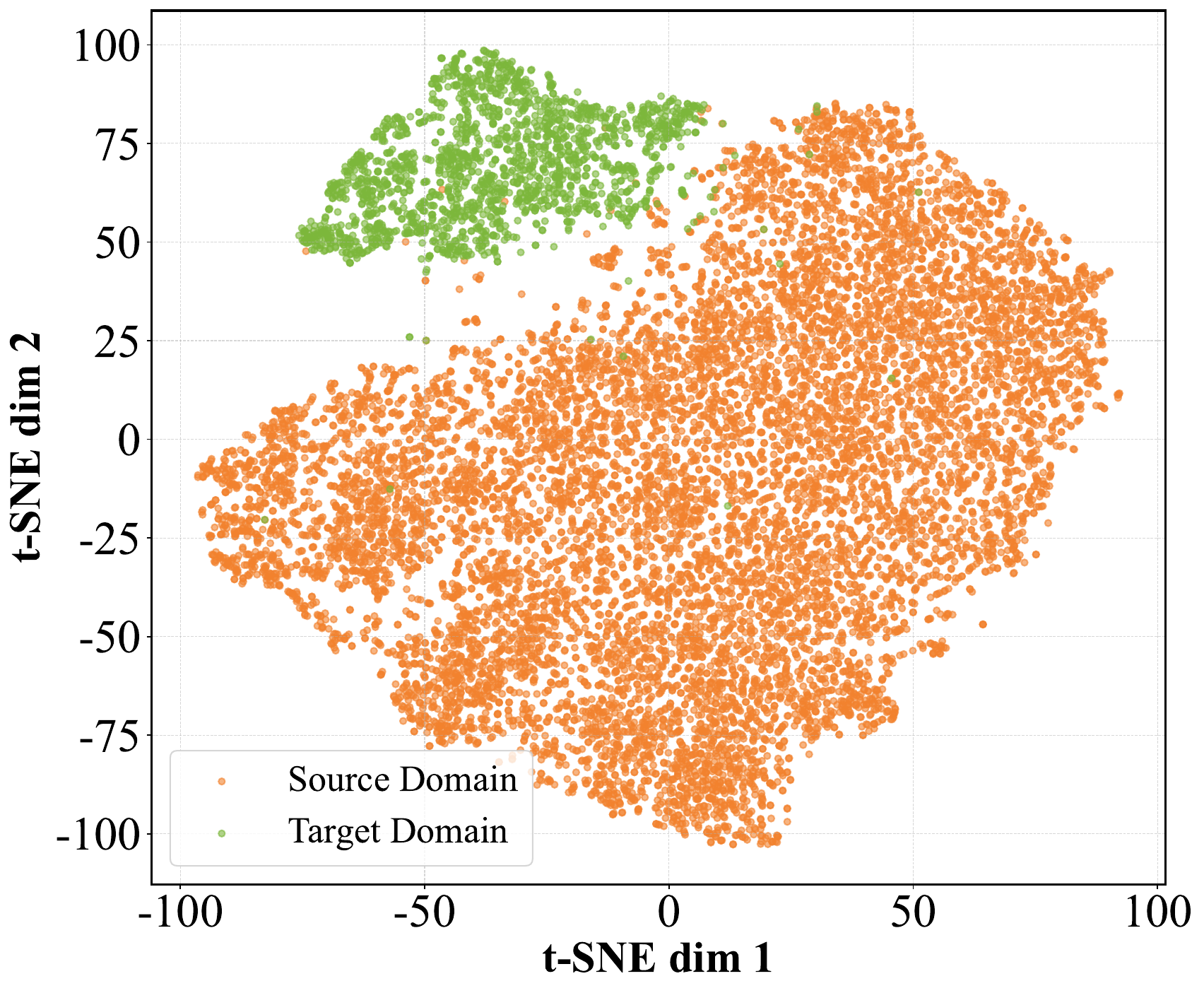} 
  \caption{t-SNE visualization of instance-level embeddings from the source and target domains.} 
  \label{fig:tsne_vis} 
\end{figure}

\begin{figure}[htbp]
  \centering 
  \includegraphics[width=\columnwidth]{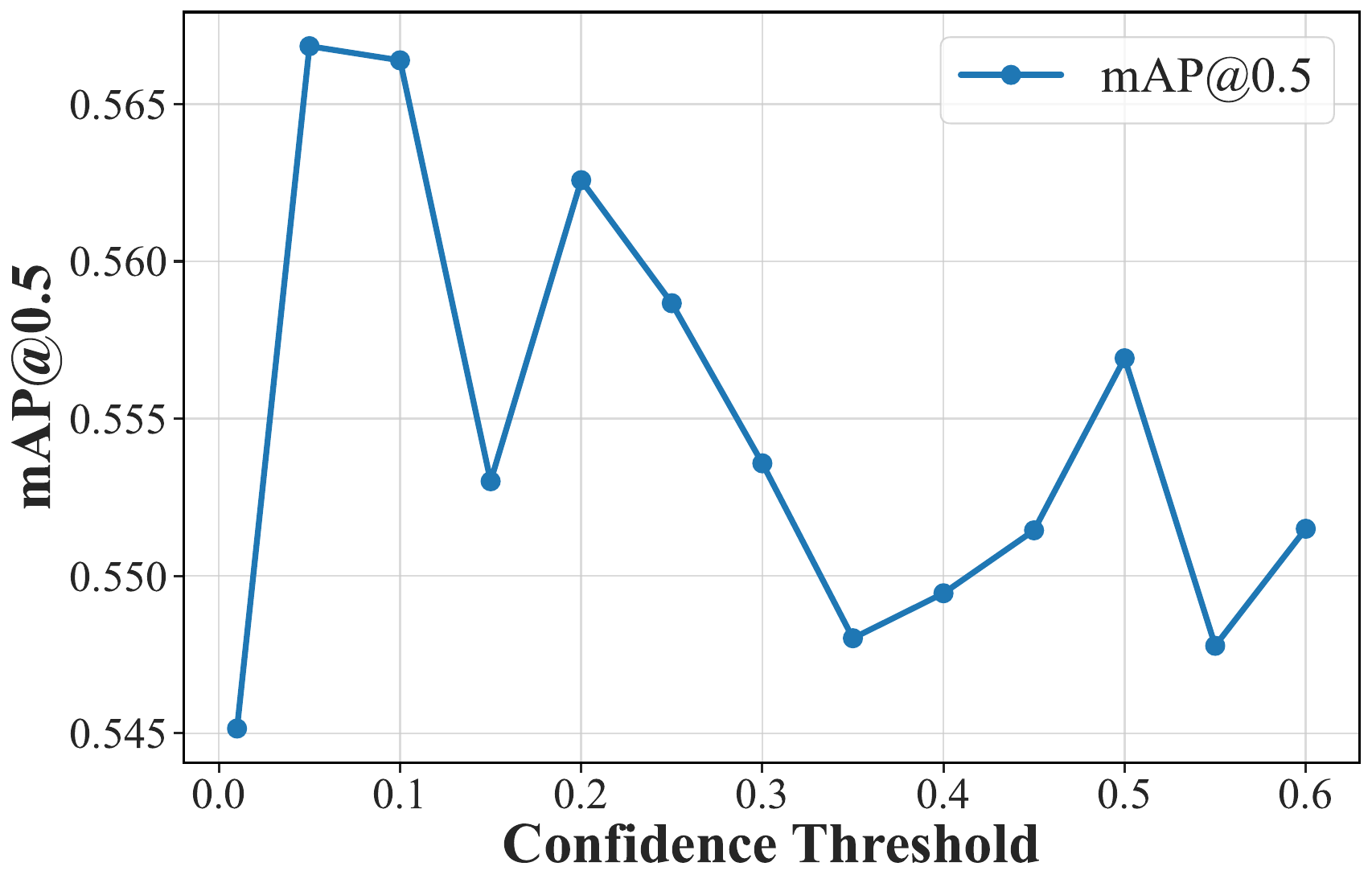} 
  \caption{The mAP@0.5 of the SimMemDA model under different initial confidence thresholds.} 
  \label{fig:confidence_map_curve} 
\end{figure}

\subsection{Qualitative Analysis}
\subsubsection{Detection Results}

In Fig. \ref{fig:detection}, we present the performance differences of various methods on the task of cross-domain SAR ship wake detection. The leftmost column shows the ground truth annotations, followed sequentially by the detection results of Source Only, ConfMix, and SimMemDA. From the Fig. \ref{fig:detection}, it can be clearly seen that these methods exhibit significant differences in detection performance. Specifically, we observe that SimMemDA has a lower missed detection rate; for instance, in the second row, SimMemDA successfully detects the ship wake. Furthermore, the detection boxes of SimMemDA can more precisely cover the boundaries of the actual targets. As shown in the first row, this method demonstrates better target localization capability, producing detection boxes with higher accuracy and greater consistency with the ground truth annotations. This indicates that the SimMemDA significantly improves the consistency between detection results and ground truth targets, effectively reducing missed detections and false positives, thus demonstrating higher accuracy and robustness in the task of cross-domain SAR wake detection.

\begin{figure}[htbp]
  \centering 
  \includegraphics[width=\columnwidth]{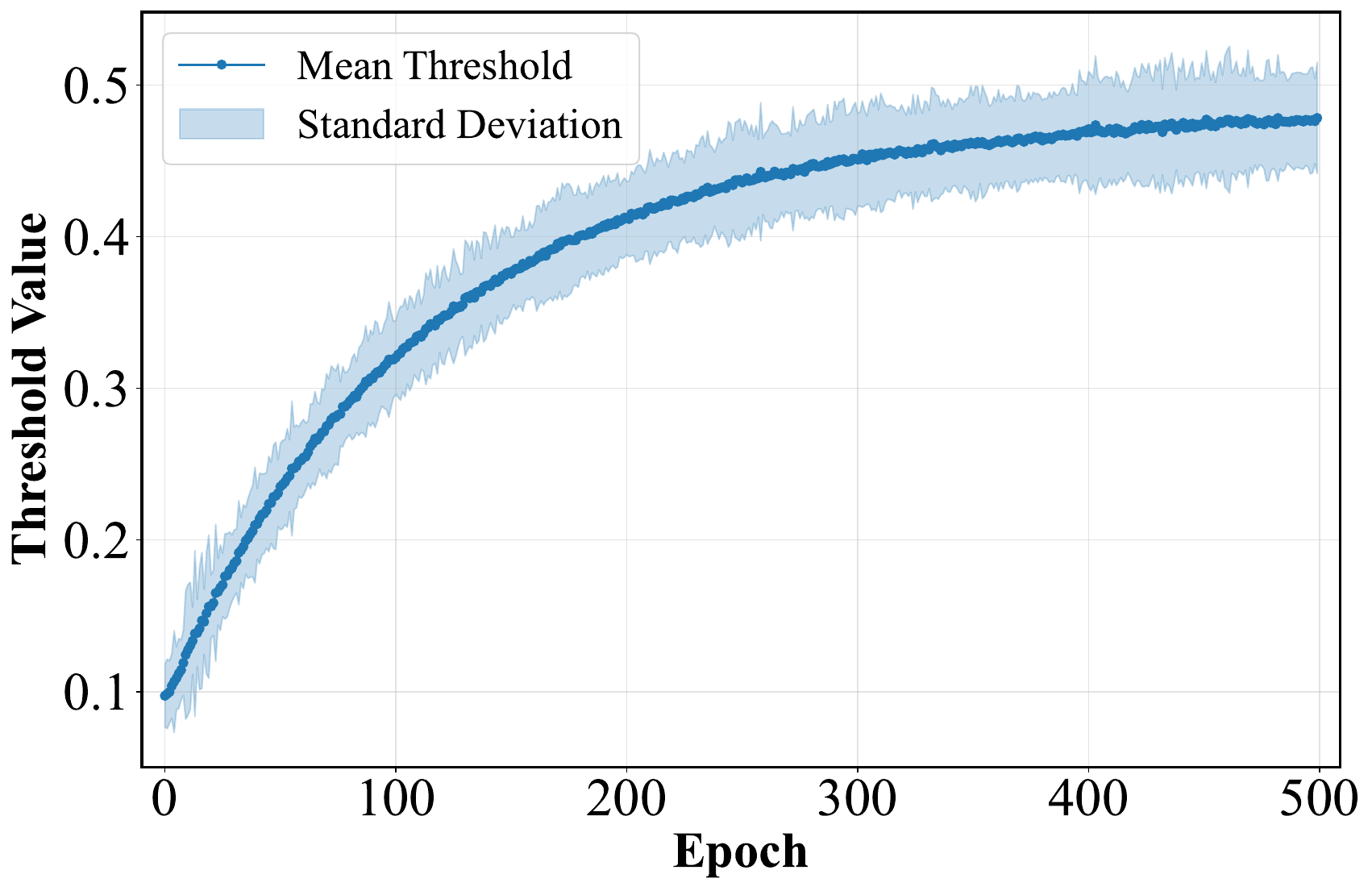} 
  \caption{The variation trends of confidence thresholds in pseudo-label calibration.} 
  \label{fig:confidence_thresh_convergence_academic} 
\end{figure}

\subsubsection{Features Visualization}
In order to analyze the properties of the features learned by SimMemDA, we present in Fig. \ref{fig:heatmap} the feature visualization results detected by Source Only and SimMemDA. Compared with the Source Only, the features in the object region detected by SimMemDA are more concentrated, more aligned with the real bounding box, and effectively suppress irrelevant interference. This indicates that our method is capable of successfully extracting the features of ship wakes in SAR images.

\begin{table}[htbp]
  \centering
  \caption{Comparison results of different discrepancy evaluation functions}
  \begin{tabular}{cccc}
    \toprule
             & \textbf{$L_2$ distance} & \textbf{$K$-means} & \textbf{GMM}   \\
    \midrule
    mAP@0.5   & \textbf{57.03}       & 54.71   & 54.32 \\
    mAP@0.5:0.05:0.95 & \textbf{19.65}       & 18.65   & 17.88 \\
    \bottomrule
  \end{tabular}
  \label{tab:estimation}
\end{table}

\subsubsection{Discrepancy Estimation Function}

To validate the effectiveness of different similarity metrics in the source domain filtering strategy, we compared the $L_2$ distance, $K$-means, and GMM discrepancy evaluation functions. As shown in Table~\ref{tab:estimation}, the performance achieved with the $L_2$ distance is the best, significantly surpassing that of $K$-means and GMM. This difference indicates that although $K$-means and GMM are capable of modeling more complex distribution structures, in the SAR wake detection task, the simple $L_2$ distance is more suitable as a similarity metric, providing more stable and effective source sample filtering. To further explain this difference, we present a t-SNE visualization of source and target domain features in Fig.~\ref{fig:tsne_vis}. From the visualization results, it can be observed that target domain samples form a relatively compact unimodal cluster in the embedding space, while source domain samples are distributed more diffusely. This unimodal distribution allows the global prototype to well represent the target domain features. Consequently, prototype-based $L_2$ distance can effectively characterize the target domain distribution and select highly similar source samples. In contrast, both $K$-means and GMM assume that the target domain distribution is multimodal, which leads to excessive partitioning of the compact target cluster and causes potentially useful samples to be incorrectly discarded, thereby degrading performance.

\begin{figure}[htbp]
  \centering 
  \includegraphics[width=\columnwidth]{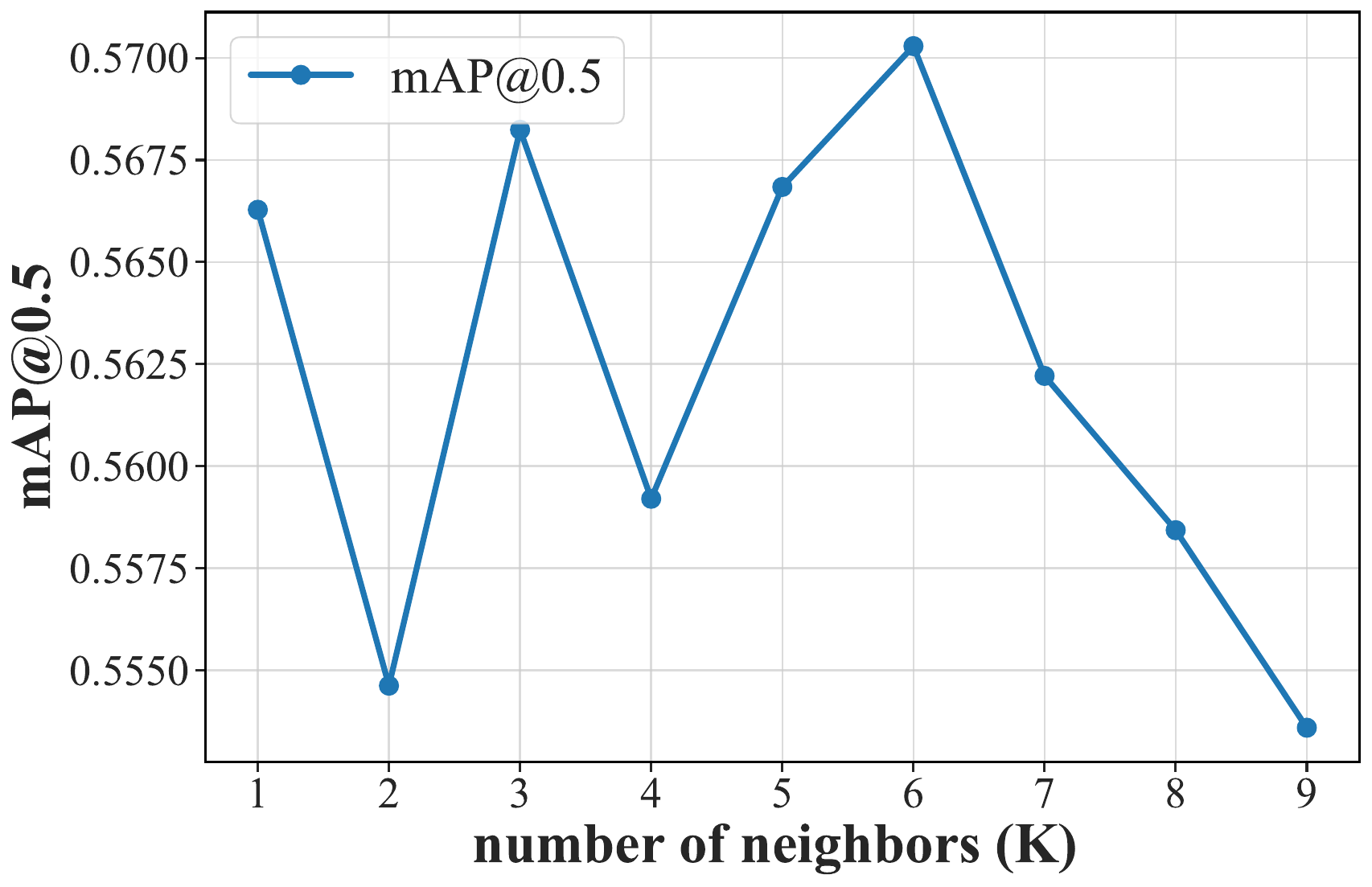} 
  \caption{Effect of the neighbor number $K$ in the feature-confidence memory bank on the detection performance.} 
  \label{fig:knn_num} 
\end{figure}

\begin{figure}[htbp]
  \centering 
  \includegraphics[width=\columnwidth]{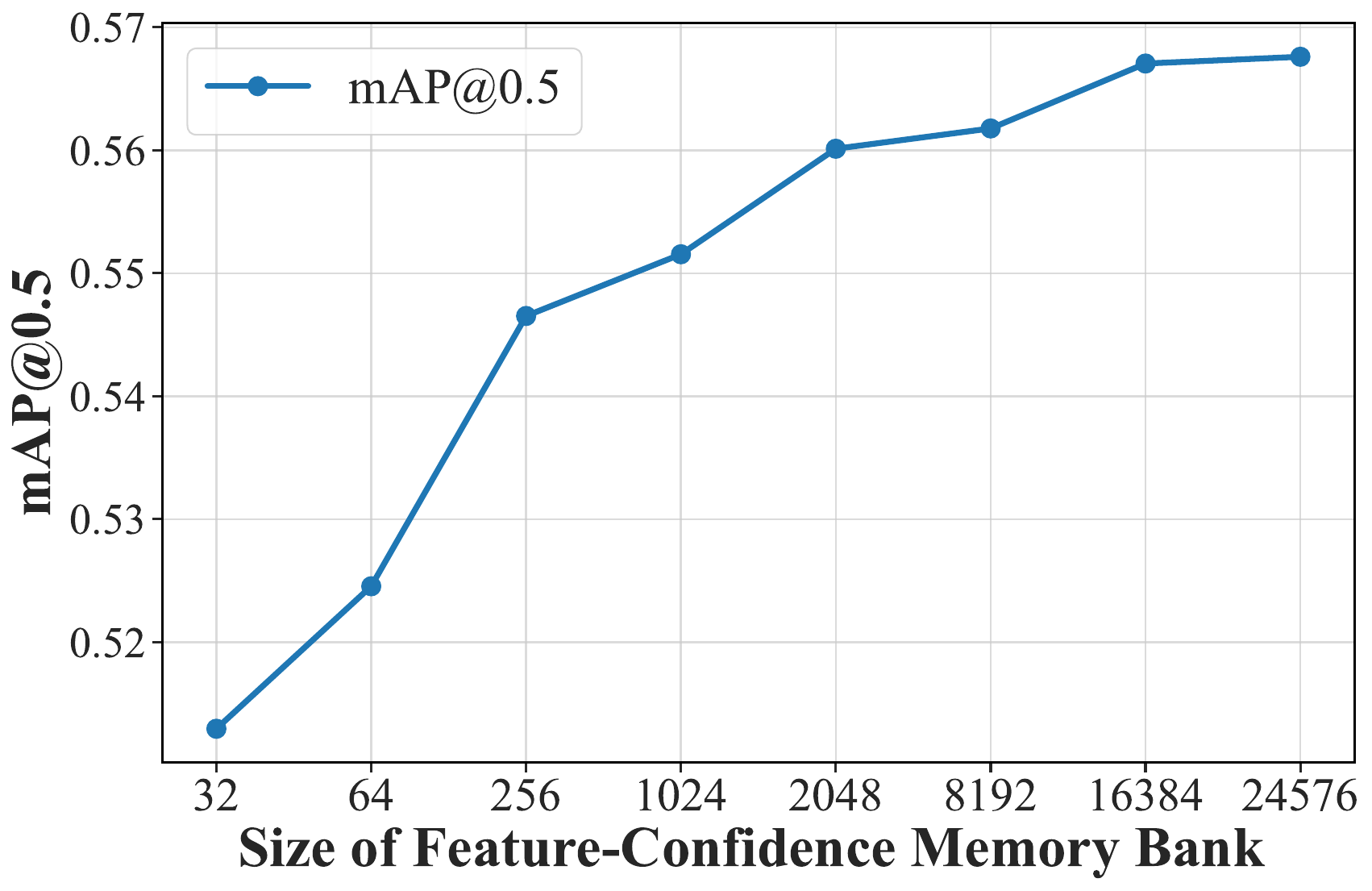} 
  \caption{Effect of feature-confidence memory bank size.} 
  \label{fig:mbsize} 
\end{figure}

\subsubsection{Confidence Threshold in Pseudo-label Calibration}
During pseudo-label calibration, the setting of the confidence threshold plays a crucial role. We experimentally investigate the impact of different initial confidence thresholds on model performance. As shown in Fig.\ref{fig:confidence_map_curve}, a lower initialization threshold tends to yield better performance. In the early training stage, a lower threshold allows the model to include more detection boxes as pseudo-labels, thereby ensuring that the detection process captures richer target information. However, if the initial confidence is set too low, excessive pseudo-labels may be introduced, which could negatively affect model training; conversely, when the initial confidence threshold is set too high, many high-quality prediction boxes that do not reach an extremely high confidence are filtered out, resulting in insufficient positive samples during training. Overall, the differences in final performance under different initial thresholds remain relatively small. This is because the proposed global-instance adaptive threshold strategy exhibits strong robustness, dynamically adjusting the filtering criteria of pseudo-labels throughout training. Fig.\ref{fig:confidence_thresh_convergence_academic} illustrates the variation of the average pseudo-label threshold during the entire training process when the initial threshold is set to 0.05. It can be observed that, under the effect of the adaptive threshold strategy, the average threshold of pseudo-labels gradually increases and eventually stabilizes at approximately 0.48. Meanwhile, its variance gradually increases during training, indicating that the threshold setting not only becomes more stringent at the global level but also adapts to individual differences among pseudo-labels. In other words, this strategy adaptively assigns the most appropriate threshold to each pseudo-label, thereby ensuring the reliability of high-confidence pseudo-labels while maintaining the diversity and completeness of pseudo-label information.

\subsubsection{Ablation on the neighborhood number $K$ in pseudo-label calibration}

In the memory-guided geometric-aware pseudo-label calibration strategy, the number of neighbors $K$ determines the sample size participating in confidence mixing within the feature-confidence memory bank. To investigate the impact of the neighbor number on the performance of SimMemDA, we conducted experiments with $K = \{1,2, \ldots, 8,9\}$, and the results are shown in Fig.~X. It can be observed that the model performance exhibits significant fluctuations with varying $K$. When $K$ is too small the calibration relies excessively on a single neighbor and is susceptible to noise interference; when $K$ is too large, the inclusion of too many heterogeneous samples leads to confidence averaging, thereby weakening local discriminability. In contrast, a moderate number of neighbors $K$ achieves a favorable balance between leveraging contextual information and preserving local consistency, thus yielding the best detection performance. These results validate that neighborhood consistency calibration plays a crucial role in improving pseudo-label quality and enhancing model robustness.

\subsubsection{Ablation on Feature-Confidence Memory Bank Size}

In the entire method, the size of the feature-confidence memory bank plays a crucial role. To this end, we conducted experiments with different memory bank sizes, and the results are shown in Fig. \ref{fig:mbsize}. As can be observed from the figure, when the memory bank size is small, due to the limited number of domain samples available for reference, the pseudo-label calibration process relies excessively on a single or a few samples, making it difficult to fully exploit the effect of calibration, which leads to relatively low detection performance. As the memory bank size gradually increases, the number of samples contained in the memory bank increases significantly, enabling the model to more effectively suppress noisy pseudo-labels during confidence fusion and geometric constraints, thereby improving the stability and reliability of the calibration. This phenomenon indicates that a larger memory bank can provide richer neighborhood consistency information for pseudo-labels and complement the geometry-aware calibration mechanism, thereby significantly enhancing the discriminative power and robustness of pseudo-labels.

\section{Conclusion}\label{section_5}
This paper addresses the problem of cross-domain unsupervised domain adaptive wake detection from optical to SAR imagery. We propose an unsupervised domain adaptation framework, SimMemDA, which leverages feature similarity guidance and feature memory guidance, constructing a multi-stage collaborative strategy spanning input-level alignment, sample filtering, pseudo-label calibration, and hybrid training. At the input level, we design a structure-preserving style transfer model, WakeGAN, which is constrained by a Frequency Selection Unit, a Detail Enhancement Guide, and a Structure Preserving Guide. Together with spectral preservation and cyclic spectral consistency losses, it achieves dual fidelity of wake geometry and scattering texture, significantly mitigating cross-domain discrepancies between optical and SAR images. In the sample filtering stage, we introduce a similarity-guided source domain filtering mechanism that effectively removes source samples with large discrepancies from the target domain, thereby avoiding the accumulation of negative transfer. In the pseudo-label generation and calibration stage, we propose a memory-guided geometry-aware confidence calibration strategy. By aggregating neighborhood information via a feature-confidence memory bank and incorporating wake geometry priors, this strategy realizes dynamic calibration of pseudo-label confidence and adaptive threshold-based selection, significantly enhancing pseudo-label reliability. Furthermore, by leveraging hybrid-domain samples under the constraint of a consistency loss, the model gradually adapts to the target distribution, thereby strengthening the learning of target domain features and enabling the detector to achieve superior generalization in SAR scenarios. SimMemDA effectively exploits the rich annotations in optical imagery to enhance SAR wake detection performance in the absence of labeled SAR data, demonstrating strong robustness across different resolutions and imaging conditions. Future work will systematically evaluate the impact of SAR wake characteristics under different band and polarization conditions on cross-domain wake detection, and further examine the adaptability of the method in complex sea states, extreme weather, and multi-modal scenarios; meanwhile, large-scale validation and transfer extension will be conducted for broader applications in ocean monitoring and maritime defense.

\appendices

\ifCLASSOPTIONcaptionsoff
  \newpage
\fi

\bibliographystyle{IEEEtran}
\bibliography{IEEEabrv,./main}

\begin{thebibliography}{10}
\providecommand{\url}[1]{#1}
\csname url@samestyle\endcsname
\providecommand{\newblock}{\relax}
\providecommand{\bibinfo}[2]{#2}
\providecommand{\BIBentrySTDinterwordspacing}{\spaceskip=0pt\relax}
\providecommand{\BIBentryALTinterwordstretchfactor}{4}
\providecommand{\BIBentryALTinterwordspacing}{\spaceskip=\fontdimen2\font plus
\BIBentryALTinterwordstretchfactor\fontdimen3\font minus \fontdimen4\font\relax}
\providecommand{\BIBforeignlanguage}[2]{{%
\expandafter\ifx\csname l@#1\endcsname\relax
\typeout{** WARNING: IEEEtran.bst: No hyphenation pattern has been}%
\typeout{** loaded for the language `#1'. Using the pattern for}%
\typeout{** the default language instead.}%
\else
\language=\csname l@#1\endcsname
\fi
#2}}
\providecommand{\BIBdecl}{\relax}
\BIBdecl

\bibitem{ding2023towards}
K.~Ding, J.~Yang, H.~Lin, Z.~Wang, D.~Wang, X.~Wang, K.~Ni, and Q.~Zhou, ``Towards real-time detection of ships and wakes with lightweight deep learning model in gaofen-3 sar images,'' \emph{Remote Sensing of Environment}, vol. 284, p. 113345, 2023.

\bibitem{guan2023method}
Y.~Guan, H.~Xu, and C.~Li, ``A method of ship wake detection in sar images based on reconstruction features and anomaly detector,'' in \emph{IGARSS 2023-2023 IEEE International Geoscience and Remote Sensing Symposium}.\hskip 1em plus 0.5em minus 0.4em\relax IEEE, 2023, pp. 6398--6401.

\bibitem{xin2023ship}
J.~Xin, J.~Gu, J.~Yang, Y.~Wen, and D.~Ding, ``Ship wake detection in sar images based on lightweight network,'' in \emph{2023 International Applied Computational Electromagnetics Society Symposium (ACES-China)}.\hskip 1em plus 0.5em minus 0.4em\relax IEEE, 2023, pp. 1--3.

\bibitem{xu2024wake2wake}
C.~Xu, Q.~Wang, X.~Wang, X.~Chao, and B.~Pan, ``Wake2wake: Feature-guided self-supervised wave suppression method for sar ship wake detection,'' \emph{IEEE Transactions on Geoscience and Remote Sensing}, 2024.

\bibitem{vesecky1982observation}
J.~F. Vesecky and R.~H. Stewart, ``The observation of ocean surface phenomena using imagery from the seasat synthetic aperture radar: An assessment,'' \emph{Journal of Geophysical Research: Oceans}, vol.~87, no.~C5, pp. 3397--3430, 1982.

\bibitem{biondi2018polarimetric}
F.~Biondi, ``A polarimetric extension of low-rank plus sparse decomposition and radon transform for ship wake detection in synthetic aperture radar images,'' \emph{IEEE Geoscience and Remote Sensing Letters}, vol.~16, no.~1, pp. 75--79, 2018.

\bibitem{jiaqiu2011novel}
A.~Jiaqiu, Q.~Xiangyang, Y.~Weidong, D.~Yunkai, L.~Fan, S.~Li, and J.~Yafei, ``A novel ship wake cfar detection algorithm based on scr enhancement and normalized hough transform,'' \emph{IEEE geoscience and remote sensing letters}, vol.~8, no.~4, pp. 681--685, 2011.

\bibitem{kang2019ship}
K.-m. Kang and D.-j. Kim, ``Ship velocity estimation from ship wakes detected using convolutional neural networks,'' \emph{IEEE Journal of Selected Topics in Applied Earth Observations and Remote Sensing}, vol.~12, no.~11, pp. 4379--4388, 2019.

\bibitem{liu2018ship}
Y.~Liu and R.~Deng, ``Ship wakes in optical images,'' \emph{Journal of Atmospheric and Oceanic Technology}, vol.~35, no.~8, pp. 1633--1648, 2018.

\bibitem{chen2018domain}
Y.~Chen, W.~Li, C.~Sakaridis, D.~Dai, and L.~Van~Gool, ``Domain adaptive faster r-cnn for object detection in the wild,'' in \emph{Proceedings of the IEEE conference on computer vision and pattern recognition}, 2018, pp. 3339--3348.

\bibitem{hsu2020every}
C.-C. Hsu, Y.-H. Tsai, Y.-Y. Lin, and M.-H. Yang, ``Every pixel matters: Center-aware feature alignment for domain adaptive object detector,'' in \emph{Computer Vision--ECCV 2020: 16th European Conference, Glasgow, UK, August 23--28, 2020, Proceedings, Part IX 16}.\hskip 1em plus 0.5em minus 0.4em\relax Springer, 2020, pp. 733--748.

\bibitem{li2022sigma}
W.~Li, X.~Liu, and Y.~Yuan, ``Sigma: Semantic-complete graph matching for domain adaptive object detection,'' in \emph{Proceedings of the IEEE/CVF conference on computer vision and pattern recognition}, 2022, pp. 5291--5300.

\bibitem{zhao2022task}
L.~Zhao and L.~Wang, ``Task-specific inconsistency alignment for domain adaptive object detection,'' in \emph{Proceedings of the IEEE/CVF conference on computer vision and pattern recognition}, 2022, pp. 14\,217--14\,226.

\bibitem{li2022scan++}
W.~Li, X.~Liu, and Y.~Yuan, ``Scan++: Enhanced semantic conditioned adaptation for domain adaptive object detection,'' \emph{IEEE Transactions on Multimedia}, vol.~25, pp. 7051--7061, 2022.

\bibitem{li2022scan}
W.~Li, X.~Liu, X.~Yao, and Y.~Yuan, ``Scan: Cross domain object detection with semantic conditioned adaptation,'' in \emph{Proceedings of the AAAI Conference on Artificial Intelligence}, vol.~36, no.~2, 2022, pp. 1421--1428.

\bibitem{li2021category}
S.~Li, J.~Huang, X.-S. Hua, and L.~Zhang, ``Category dictionary guided unsupervised domain adaptation for object detection,'' in \emph{Proceedings of the AAAI conference on artificial intelligence}, vol.~35, no.~3, 2021, pp. 1949--1957.

\bibitem{ramamonjison2021simrod}
R.~Ramamonjison, A.~Banitalebi-Dehkordi, X.~Kang, X.~Bai, and Y.~Zhang, ``Simrod: A simple adaptation method for robust object detection,'' in \emph{Proceedings of the IEEE/CVF International Conference on Computer Vision}, 2021, pp. 3570--3579.

\bibitem{mattolin2023confmix}
G.~Mattolin, L.~Zanella, E.~Ricci, and Y.~Wang, ``Confmix: Unsupervised domain adaptation for object detection via confidence-based mixing,'' in \emph{Proceedings of the IEEE/CVF Winter Conference on Applications of Computer Vision}, 2023, pp. 423--433.

\bibitem{yang2025versatile}
R.~Yang, T.~Tian, and J.~Tian, ``Versatile teacher: A class-aware teacher--student framework for cross-domain adaptation,'' \emph{Pattern Recognition}, vol. 158, p. 111024, 2025.

\bibitem{chen2022learning}
M.~Chen, W.~Chen, S.~Yang, J.~Song, X.~Wang, L.~Zhang, Y.~Yan, D.~Qi, Y.~Zhuang, D.~Xie \emph{et~al.}, ``Learning domain adaptive object detection with probabilistic teacher,'' in \emph{International Conference on Machine Learning}.\hskip 1em plus 0.5em minus 0.4em\relax PMLR, 2022, pp. 3040--3055.

\bibitem{he2022cross}
M.~He, Y.~Wang, J.~Wu, Y.~Wang, H.~Li, B.~Li, W.~Gan, W.~Wu, and Y.~Qiao, ``Cross domain object detection by target-perceived dual branch distillation,'' in \emph{Proceedings of the IEEE/CVF Conference on Computer Vision and Pattern Recognition}, 2022, pp. 9570--9580.

\bibitem{xu2020exploring}
C.-D. Xu, X.-R. Zhao, X.~Jin, and X.-S. Wei, ``Exploring categorical regularization for domain adaptive object detection,'' in \emph{Proceedings of the IEEE/CVF conference on computer vision and pattern recognition}, 2020, pp. 11\,724--11\,733.

\bibitem{cao2023contrastive}
S.~Cao, D.~Joshi, L.-Y. Gui, and Y.-X. Wang, ``Contrastive mean teacher for domain adaptive object detectors,'' in \emph{Proceedings of the IEEE/CVF conference on computer vision and pattern recognition}, 2023, pp. 23\,839--23\,848.

\bibitem{kim2019diversify}
T.~Kim, M.~Jeong, S.~Kim, S.~Choi, and C.~Kim, ``Diversify and match: A domain adaptive representation learning paradigm for object detection,'' in \emph{Proceedings of the IEEE/CVF Conference on Computer Vision and Pattern Recognition}, 2019, pp. 12\,456--12\,465.

\bibitem{wang2021afan}
H.~Wang, S.~Liao, and L.~Shao, ``Afan: Augmented feature alignment network for cross-domain object detection,'' \emph{IEEE Transactions on Image Processing}, vol.~30, pp. 4046--4056, 2021.

\bibitem{shi2022unsupervised}
Y.~Shi, L.~Du, Y.~Guo, and Y.~Du, ``Unsupervised domain adaptation based on progressive transfer for ship detection: From optical to sar images,'' \emph{IEEE Transactions on Geoscience and Remote Sensing}, vol.~60, pp. 1--17, 2022.

\bibitem{xi2024cromoda}
C.~Xi, Z.~Wang, W.~Wang, X.~Xie, J.~Kang, and R.~Fernandez-Beltran, ``Cromoda: Unsupervised oriented sar ship detection via cross-modality distribution alignment,'' \emph{IEEE Journal of Selected Topics in Applied Earth Observations and Remote Sensing}, 2024.

\bibitem{chan2024complex}
H.~Chan, X.~Qiu, X.~Gao, and D.~Lu, ``A complex background sar ship target detection method based on fusion tensor and cross-domain adversarial learning,'' \emph{Remote Sensing}, vol.~16, no.~18, p. 3492, 2024.

\bibitem{zhou2024domain}
Z.~Zhou, L.~Zhao, K.~Ji, and G.~Kuang, ``A domain adaptive few-shot sar ship detection algorithm driven by the latent similarity between optical and sar images,'' \emph{IEEE Transactions on Geoscience and Remote Sensing}, 2024.

\bibitem{beylkin1985imaging}
G.~Beylkin, ``Imaging of discontinuities in the inverse scattering problem by inversion of a causal generalized radon transform,'' \emph{Journal of mathematical physics}, vol.~26, no.~1, pp. 99--108, 1985.

\bibitem{ballard1981generalizing}
D.~H. Ballard, ``Generalizing the hough transform to detect arbitrary shapes,'' \emph{Pattern recognition}, vol.~13, no.~2, pp. 111--122, 1981.

\bibitem{murphy1986linear}
L.~M. Murphy, ``Linear feature detection and enhancement in noisy images via the radon transform,'' \emph{Pattern recognition letters}, vol.~4, no.~4, pp. 279--284, 1986.

\bibitem{rey1990application}
M.~T. Rey, J.~K. Tunaley, J.~Folinsbee, P.~A. Jahans, J.~Dixon, and M.~R. Vant, ``Application of radon transform techniques to wake detection in seasat-a sar images,'' \emph{IEEE Transactions on Geoscience and Remote Sensing}, vol.~28, no.~4, pp. 553--560, 1990.

\bibitem{biondi2017low}
F.~Biondi, ``Low-rank plus sparse decomposition and localized radon transform for ship-wake detection in synthetic aperture radar images,'' \emph{IEEE Geoscience and Remote Sensing Letters}, vol.~15, no.~1, pp. 117--121, 2017.

\bibitem{srivastava2022analysis}
R.~Srivastava and J.~Christmas, ``Analysis of sea waves and ship wake detection,'' in \emph{OCEANS 2022-Chennai}.\hskip 1em plus 0.5em minus 0.4em\relax IEEE, 2022, pp. 1--10.

\bibitem{hong2022target}
J.~Hong, X.~Liu, and H.~Dong, ``Target recognition method based on ship wake extraction from remote sensing image,'' in \emph{Asian Simulation Conference}.\hskip 1em plus 0.5em minus 0.4em\relax Springer, 2022, pp. 214--226.

\bibitem{jiang2024ship}
Y.~Jiang, K.~Li, Z.~Yang, and T.~Liu, ``Ship wake detection based on polarimetric enhancement and deep learning via a simulated full-polarized dataset d{\'e}tection des sillages de navires bas{\'e}e sur l’am{\'e}lioration polarim{\'e}trique et l’apprentissage profond via un ensemble de donn{\'e}es simul{\'e}es enti{\`e}rement polaris{\'e}es,'' \emph{IEEE Canadian Journal of Electrical and Computer Engineering}, 2024.

\bibitem{liu2021novel}
Y.~Liu, J.~Zhao, and Y.~Qin, ``A novel technique for ship wake detection from optical images,'' \emph{Remote Sensing of Environment}, vol. 258, p. 112375, 2021.

\bibitem{liu2024kelvin}
Y.~Liu and J.~Zhao, ``Kelvin wake detection from large-scale optical imagery using simulated data trained deep neural network,'' \emph{Ocean Engineering}, vol. 297, p. 117075, 2024.

\bibitem{he2023cross}
J.~He, N.~Su, C.~Xu, Y.~Liao, Y.~Yan, C.~Zhao, W.~Hou, and S.~Feng, ``A cross-modality feature transfer method for target detection in sar images,'' \emph{IEEE Transactions on Geoscience and Remote Sensing}, vol.~61, pp. 1--15, 2023.

\bibitem{yuan2023adaptive}
Y.~Yuan, Z.~Rao, C.~Lin, Y.~Huang, and X.~Ding, ``Adaptive ship detection from optical to sar images,'' \emph{IEEE Geoscience and Remote Sensing Letters}, vol.~20, pp. 1--5, 2023.

\bibitem{liu2024confidence}
C.~Liu, Y.~Dong, Y.~Zhang, and X.~Li, ``Confidence-driven region mixing for optical remote sensing domain adaptation object detection,'' \emph{IEEE Transactions on Geoscience and Remote Sensing}, 2024.

\bibitem{shi2024unsupervised}
Y.~Shi, Y.~Li, L.~Du, Y.~Du, and Y.~Guo, ``Unsupervised domain adaptative sar target detection based on feature decomposition and uncertainty-guided self-training,'' \emph{IEEE Journal of Selected Topics in Applied Earth Observations and Remote Sensing}, 2024.

\bibitem{yang2024unsupervised}
Y.~Yang, J.~Chen, L.~Sun, Z.~Zhou, Z.~Huang, and B.~Wu, ``Unsupervised domain-adaptive sar ship detection based on cross-domain feature interaction and data contribution balance,'' \emph{Remote Sensing}, vol.~16, no.~2, p. 420, 2024.

\bibitem{luo2024sar}
C.~Luo, Y.~Zhang, J.~Guo, Y.~Hu, G.~Zhou, H.~You, and X.~Ning, ``Sar-cdss: A semi-supervised cross-domain object detection from optical to sar domain,'' \emph{Remote Sensing}, vol.~16, no.~6, p. 940, 2024.

\bibitem{huang2024domain}
H.~Huang, J.~Guo, H.~Lin, Y.~Huang, and X.~Ding, ``Domain adaptive oriented object detection from optical to sar images,'' \emph{IEEE Transactions on Geoscience and Remote Sensing}, 2024.

\bibitem{chen2022pixel}
Z.~Chen, L.~Zhao, Q.~He, and G.~Kuang, ``Pixel-level and feature-level domain adaptation for heterogeneous sar target recognition,'' \emph{IEEE Geoscience and Remote Sensing Letters}, vol.~19, pp. 1--5, 2022.

\bibitem{pu2023ship}
X.~Pu, H.~Jia, Y.~Xin, F.~Wang, and H.~Wang, ``Ship detection in low-quality sar images via an unsupervised domain adaption method,'' \emph{Remote Sensing}, vol.~15, no.~13, p. 3326, 2023.

\bibitem{zhu2017unpaired}
J.-Y. Zhu, T.~Park, P.~Isola, and A.~A. Efros, ``Unpaired image-to-image translation using cycle-consistent adversarial networks,'' in \emph{Proceedings of the IEEE international conference on computer vision}, 2017, pp. 2223--2232.

\bibitem{liu2021swin}
Z.~Liu, Y.~Lin, Y.~Cao, H.~Hu, Y.~Wei, Z.~Zhang, S.~Lin, and B.~Guo, ``Swin transformer: Hierarchical vision transformer using shifted windows,'' in \emph{Proceedings of the IEEE/CVF international conference on computer vision}, 2021, pp. 10\,012--10\,022.

\bibitem{li2022robust}
S.~Li, X.~Lv, J.~Ren, and J.~Li, ``A robust 3d density descriptor based on histogram of oriented primary edge structure for sar and optical image co-registration,'' \emph{Remote Sensing}, vol.~14, no.~3, p. 630, 2022.

\bibitem{dempster1977maximum}
A.~P. Dempster, N.~M. Laird, and D.~B. Rubin, ``Maximum likelihood from incomplete data via the em algorithm,'' \emph{Journal of the royal statistical society: series B (methodological)}, vol.~39, no.~1, pp. 1--22, 1977.

\bibitem{he2020momentum}
K.~He, H.~Fan, Y.~Wu, S.~Xie, and R.~Girshick, ``Momentum contrast for unsupervised visual representation learning,'' in \emph{Proceedings of the IEEE/CVF conference on computer vision and pattern recognition}, 2020, pp. 9729--9738.

\bibitem{wangfreematch}
Y.~Wang, H.~Chen, Q.~Heng, W.~Hou, Y.~Fan, Z.~Wu, J.~Wang, M.~Savvides, T.~Shinozaki, B.~Raj \emph{et~al.}, ``Freematch: Self-adaptive thresholding for semi-supervised learning,'' in \emph{The Eleventh International Conference on Learning Representations}.

\bibitem{belal2024multi}
A.~Belal, A.~Meethal, F.~P. Romero, M.~Pedersoli, and E.~Granger, ``Multi-source domain adaptation for object detection with prototype-based mean teacher,'' in \emph{Proceedings of the IEEE/CVF winter conference on applications of computer vision}, 2024, pp. 1277--1286.

\bibitem{weng2024mean}
W.~Weng and C.~Yuan, ``Mean teacher detr with masked feature alignment: A robust domain adaptive detection transformer framework,'' in \emph{Proceedings of the AAAI Conference on Artificial Intelligence}, vol.~38, no.~6, 2024, pp. 5912--5920.

\bibitem{xue2021rethinking}
F.~Xue, W.~Jin, S.~Qiu, and J.~Yang, ``Rethinking automatic ship wake detection: state-of-the-art cnn-based wake detection via optical images,'' \emph{IEEE Transactions on Geoscience and Remote Sensing}, vol.~60, pp. 1--22, 2021.

\bibitem{xu2024opensarwake}
C.~Xu and X.~Wang, ``Opensarwake: A large-scale sar dataset for ship wake recognition with a feature refinement oriented detector,'' \emph{IEEE Geoscience and Remote Sensing Letters}, 2024.

\bibitem{liu2021adaattn}
S.~Liu, T.~Lin, D.~He, F.~Li, M.~Wang, X.~Li, Z.~Sun, Q.~Li, and E.~Ding, ``Adaattn: Revisit attention mechanism in arbitrary neural style transfer,'' in \emph{Proceedings of the IEEE/CVF international conference on computer vision}, 2021, pp. 6649--6658.

\bibitem{kim2024unpaired}
B.~Kim, G.~Kwon, K.~Kim, and J.~C. Ye, ``Unpaired image-to-image translation via neural schr{\"o}dinger bridge,'' in \emph{12th International Conference on Learning Representations, ICLR 2024}, 2024.

\bibitem{mackiewicz1993principal}
A.~Ma{\'c}kiewicz and W.~Ratajczak, ``Principal components analysis (pca),'' \emph{Computers \& Geosciences}, vol.~19, no.~3, pp. 303--342, 1993.

\end{thebibliography}

\begin{IEEEbiography}[{\includegraphics[width=1in,height=1.25in,clip,keepaspectratio]{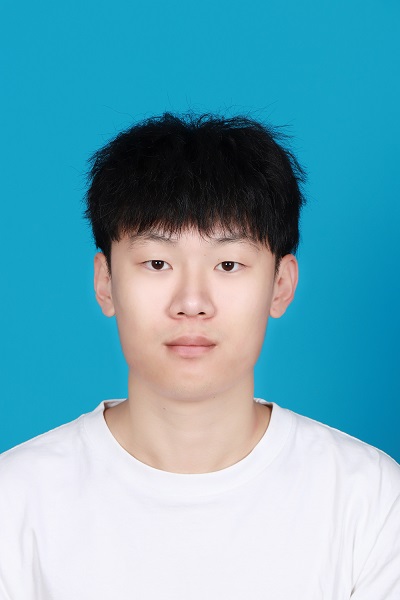}}]{He Gao} received the B.S. degree in computer science and technology from the Qilu University of Technology, Jinan, China, in 2019, and the M.S. degree in computer technology from the College of Computer Science and Technology, Qingdao University, Qingdao, China, in 2023. He is currently pursuing the Ph.D. degree at Qingdao University.

His research interests include deep learning, transfer learning, and artificial intelligence oceanography.
\end{IEEEbiography}

\begin{IEEEbiography}[{\includegraphics[width=1in,height=1.25in,clip,keepaspectratio]{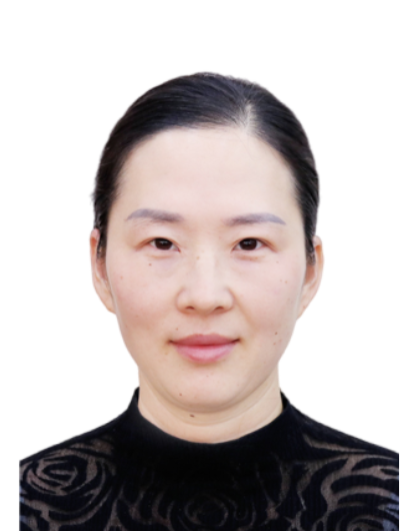}}]{Baoxiang Huang}
(Senior Member, IEEE) received the B.S. degree in traffic engineering from the Shandong University of Technology, Zibo, China, in 2002, the M.S. degree in mechatronic engineering from Shandong University, Jinan, China, in 2005, and the Ph.D. degree in computer engineering from the Ocean University of China, Qingdao, China, in 2011. 

She is currently a Professor with the College of
Computer Science and Technology, Qingdao University, Qingdao, China. She was an Academic Visitor of Nottingham University, Nottingham, U.K. Her research interests include remote sensing image processing and analysis, big data oceanography, and artificial intelligence.
\end{IEEEbiography}

\begin{IEEEbiography}[{\includegraphics[width=1in,height=1.25in,clip,keepaspectratio]{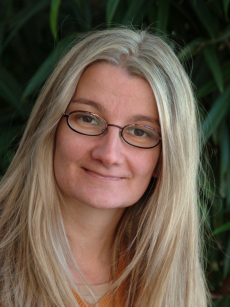}}]{Milena Radenkovic}
  (Member, IEEE) received the Dipl.-Ing. degree in electric and electronic engineering from the University of Nis, Nis, Serbia, in 1998, and the Ph.D. degree in computer science from The University of Nottingham, Nottingham, U.K., in 2002.
  
  She has authored more than 80 articles in premium conferences and journal venues. Her research interests include intelligent mobile and disconnectiontolerant networking, complex temporal graphs, selforganized security, distributed predictive analytics with applications to autonomous vehicles, mobile social networks, smart manufacturing, and predictive telemetry.
  
  Dr. Radenkovic was a recipient of Multiple EPSRC and EU grants for her research. She has organized and chaired multiple ACM and IEEE conferences and served on many program committees. She is an Editor for premium journals, such as the Ad Hoc Networks (Elsevier), IEEE TRANSACTIONS ON PARALLEL AND DISTRIBUTED COMPUTING, and ACM Multimedia.   
  \end{IEEEbiography}

\begin{IEEEbiography}[{\includegraphics[width=1in,height=1.25in,clip,keepaspectratio]{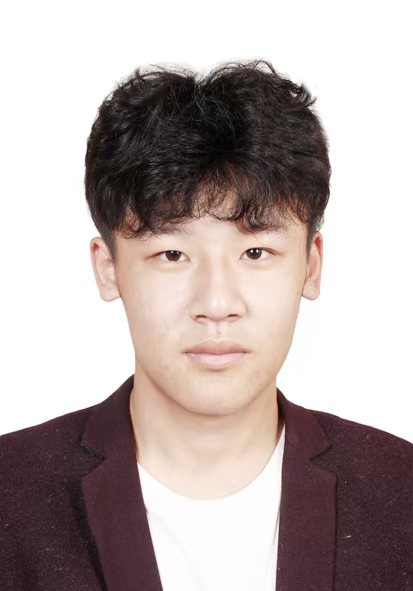}}]{Borui Li}
is pursuing the M.S. degree in computer application technology from the Qingdao University, Qingdao. His research interests include satellite remote sensing of the ocean, object detection, and deep learning.    
\end{IEEEbiography}

\begin{IEEEbiography}[{\includegraphics[width=1in,height=1.25in,clip,keepaspectratio]{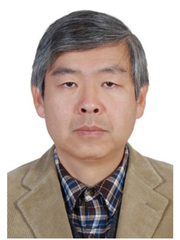}}]{Ge Chen} (Senior Member, IEEE) received the B.S. degree in marine physics, the M.S. degree in satellite oceanography, and the Ph.D. degree in physical
oceanography from the Ocean University of China (OUC), Qingdao, China, in 1988, 1990, and 1993, respectively. 

He was a Postdoctoral Fellow with The French Research Institute for the Exploitation of the Sea (IFREMER), Brest, France, from 1994 to 1996. Since 1997, he has been a Professor of Satellite Oceanography and Meteorology with OUC. He is currently the Deputy
Dean with the Institute for Advanced Marine Sciences, OUC, and the Chief Scientist for Ocean Science Satellite Missions with the National Laboratory of Ocean Science and Technology, Qingdao. He has authored or coauthored more than 110 peer-reviewed scientific articles published in internationally recognized journals. His research interests include satellite remote sensing of the ocean and big data oceanography. 

Dr. Chen was the Executive Secretary of the International Pan Ocean Remote Sensing Conference (PORSEC) Association from 1998 to 2002. He was a recipient of the National Science Fund for Outstanding Young Scientists awarded
by the Natural Science Foundation of China in 2001. He became the Chair Professor of Cheung Kong Scholars Program nominated by the Chinese Ministry
of Education.
\end{IEEEbiography}







\end{document}